\documentclass[11pt]{article}

\usepackage{fullpage,times,url,bm}

\usepackage{amsthm,amsfonts,amsmath,amssymb,epsfig,color,float,graphicx,verbatim}
\usepackage{algorithm,algorithmic}
\usepackage{bbm}
\usepackage{enumitem}

\usepackage{graphicx}
\usepackage{subcaption}

\usepackage{array}

\usepackage[dvipsnames]{xcolor}

\usepackage[colorlinks,linkcolor = RawSienna, urlcolor  = Green, citecolor = Green, anchorcolor = ForestGreen,bookmarks=False]{hyperref}

\newtheorem{theorem}{Theorem}[section]
\newtheorem{lemma}{Lemma}[section]
\newtheorem{corollary}{Corollary}[section]

\renewcommand{\eqref}[1]{Eq.~(\ref{#1})}

\usepackage{amssymb}

\usepackage{bbm}
\newcommand{\onefunc}{\mathbbm{1}}

\newcommand{\stam}[1]{}
\newcommand{\ignore}[1]{}
\usepackage{comment}

\usepackage{mathtools}

\newtheorem{assumption}[theorem]{Assumption}

\DeclareMathOperator{\poly}{poly}

\DeclareMathOperator*{\E}{\mathbb{E}}
\DeclareMathOperator{\bin}{bin}

\usepackage[style = alphabetic,citestyle=alphabetic,maxalphanames=6,maxcitenames=4,maxbibnames=99,sorting=nyt,natbib=true,backref=true]{biblatex}
\DefineBibliographyStrings{english}{%
  backrefpage = {Cited on page},
  backrefpages = {Cited on pages},
}

\makeatletter
\newcommand{\printfnsymbol}[1]{%
  \textsuperscript{\@fnsymbol{#1}}%
}
\makeatother

\usepackage[T1]{fontenc}
\usepackage{microtype,booktabs,needspace,placeins}
\setlist[enumerate]{itemsep=0.35em,topsep=0.4em}
\numberwithin{equation}{section}
\hypersetup{
  pdftitle={Hardness of Learning Juntas and Decision Trees from Subexponential LWE},
  pdfauthor={Gal Vardi},
  pdfsubject={Improper hardness of PAC learning juntas and decision trees}
}
\newcommand{\Z}{\mathbb Z}

\newcommand{\TV}{\operatorname{TV}}
\newcommand{\bits}{\{0,1\}}
\newcommand{\Unif}{\operatorname{Unif}}
\newcommand{\Enc}{\mathsf{Enc}}
\newcommand{\Dec}{\mathsf{Dec}}
\newcommand{\Eval}{\mathsf{Eval}}

\newcommand{\err}{\operatorname{err}}
\newcommand{\cL}{\mathcal L}
\newcommand{\cU}{\mathcal U}

\title{Hardness of Learning Juntas and Decision Trees\\
from Subexponential LWE}
\author{
Gal Vardi\\
Weizmann Institute of Science\\ 
\texttt{gal.vardi@weizmann.ac.il}
}

\date{}

\begin{document}

\maketitle

\begin{abstract}
We prove time lower bounds for learning juntas and decision trees under quantitative hardness assumptions on learning with errors (LWE). Our bounds hold in the distribution-free realizable PAC model, even for improper learners. Assuming that, for some fixed $\beta\in(1/2,1)$, LWE in dimension $n$ has no $2^{O(n^\beta)}$-time distinguisher, we prove that for $\kappa(N)=\Omega((\log N)^{1/\beta-1}\,(\log\log N)^2)$, neither $\kappa(N)$-juntas on $N$ variables nor decision trees with at most $2^{\kappa(N)}$ leaves are polynomial-time learnable. This gives hardness for sublogarithmic juntas and subpolynomial-size decision trees. Strengthening the LWE assumption to exclude $2^{o(n)}$-time distinguishers lowers the threshold to $\kappa(N)=\omega(\log\log N\,\log\log\log N)$.

Moreover, we obtain quantitative lower bounds under both assumptions. In particular, under our stronger assumption, we rule out $N^{o(\log N/(\log\log N)^2)}$-time learning of $(\log N)$-juntas and decision trees with at most $N$ leaves, approaching the known upper bounds within a factor $(\log\log N)^2$ in the exponent. Our results also imply hardness for polynomial-size bounded-width DNFs and intersections of sparse halfspaces, and for polynomial thresholds and $\mathbb F_2$ polynomials of bounded degree with polynomially many monomials.
\end{abstract}

\section{Introduction}\label{sec:introduction}

Learning polynomial-size Boolean decision trees and logarithmic juntas from random examples is a long-studied problem in computational learning theory~\citep{Ser25}.
A $\kappa$-junta on $N$ variables depends on at most $\kappa$ coordinates, unknown to the learner, and can be represented by a decision tree with at most $2^\kappa$ leaves. Both classes can be learned in $N^{O(\kappa)}$ time 
under any input distribution~\citep{EH89,Ser25}. Thus logarithmic juntas and decision trees with at most $N$ leaves can be learned in $N^{O(\log N)}$ time.

\citet{ABW10} proved hardness of learning logarithmic juntas, and hence also polynomial-size decision trees and DNF formulas, under a combination of two assumptions that they introduced: The first assumption is related to the planted dense subgraph problem in hypergraphs, and the second concerns the existence of local pseudorandom generators (PRGs). Still, efforts to establish such hardness results under more widely studied assumptions have continued. For example, \citet{DSS16} proved hardness of learning DNFs under the assumption that refuting a random $K$-SAT formula is hard. Despite the result of \citet{ABW10}, the hardness of learning juntas and decision trees under common assumptions is often considered an open problem in the literature. Indeed, efficient distribution-free improper learning of juntas or decision trees from random examples was mentioned as an open question by, e.g., \citet[Section~3.2]{Ser25}, \citet[Section~6.3]{DLS14}, \citet[Section~1.3]{DRWY12}, and \citet[Section~4]{DSS16}.

Under commonly used quantitative assumptions on the standard \emph{learning with errors} (LWE) problem, we establish hardness for sublogarithmic juntas and subpolynomial-size decision trees, as well as quantitative time lower bounds. These bounds hold in the distribution-free realizable PAC model, even for improper learners, whose returned functions may lie outside the target class.

\paragraph{Polynomial-time hardness.}
Suppose that, for some fixed $\beta\in(1/2,1)$, LWE in dimension $n$ cannot be distinguished from uniform
in time $2^{O(n^\beta)}$.
We prove that $\kappa(N)$-juntas and decision trees with at most $2^{\kappa(N)}$ leaves are not polynomial-time learnable whenever
\begin{equation}\label{eq:intro-beta-growth}
 \kappa(N)=\Omega\bigl((\log N)^{\frac{1}{\beta}-1}\,(\log\log N)^2\bigr)
\end{equation}
(see Theorem~\ref{thm:beta-main}\textnormal{(i)} and Corollary~\ref{cor:trees}).
The above threshold is $o(\log N)$ for every such $\beta$, giving hardness for sublogarithmic juntas and subpolynomial-size decision trees. For example, $\beta=2/3$ gives the threshold $\sqrt{\log N}\,(\log\log N)^2$.

Under the stronger assumption that no $2^{o(n)}$-time LWE 
distinguishers exist,
the threshold for both classes improves to
\begin{equation}\label{eq:intro-no-subexp-growth}
 \kappa(N)=\omega\bigl(\log\log N\,\log\log\log N\bigr)
\end{equation}
(see Theorem~\ref{thm:no-subexp}\textnormal{(i)} and Corollary~\ref{cor:trees}).

\paragraph{Quantitative lower bounds and comparison with upper bounds.}
Under the assumption excluding $2^{o(n)}$-time LWE distinguishers, 
we
rule out learning $(\log N)$-juntas and decision trees with at most $N$ leaves in time
\[
 N^{o(\log N/(\log\log N)^2)}~.
\]
This approaches the known $N^{O(\log N)}$ upper bounds within a factor $(\log\log N)^2$ in the exponent.
More generally, we show that this assumption rules out learning $\kappa$-juntas and decision trees with at most $2^{\kappa(N)}$ leaves in time $N^{o(\kappa/(\log\kappa\,\log\log N))}$ (Theorem~\ref{thm:no-subexp}\textnormal{(ii)} and Corollary~\ref{cor:tree-runtime}).
Compared with the known $N^{O(\kappa)}$ algorithm, this leaves a factor of $\log\kappa\,\log\log N$ between the exponents.
We also give a time lower bound under the $2^{O(n^\beta)}$-time LWE assumption (Theorem~\ref{thm:beta-main}\textnormal{(ii)}).

\paragraph{Implications for bounded-width DNFs and intersections of sparse halfspaces.}
Under either assumption, the junta lower bounds also imply hardness for polynomial-size bounded-width DNFs and intersections of sparse halfspaces.
For either growth threshold, \eqref{eq:intro-beta-growth} or \eqref{eq:intro-no-subexp-growth}, our corresponding lower bounds for $\kappa$-juntas apply to DNFs with at most $2^\kappa$ terms of width at most $\kappa$, and to intersections of at most $2^\kappa$ halfspaces on the hypercube that depend on at most $\kappa$ variables  (see Section~\ref{sec:dnf}).

Bounded-width DNFs where each term has width at most $\kappa$ (also called $\kappa$-DNFs) have been studied since Valiant's work~\citep{Val84}, which gave a time upper bound of $N^{O(\kappa)}$. This bound also applies to intersections of $\kappa$-sparse halfspaces on the hypercube. Thus, under our stronger LWE assumption, our lower bound is tight up to a factor of $\log\kappa\,\log\log N$ in the exponent.

Several lower bounds for improperly learning DNFs and intersections of halfspaces are known. First, as we already mentioned, \citet{ABW10} and \citet{DSS16} proved hardness of learning DNFs under different assumptions. The latter result holds already for DNFs with $\omega(\log N)$ terms. \citet{DV21} proved hardness already for DNFs with $\omega(1)$ terms, assuming the existence of local PRGs with polynomial stretch. \citet{ABCM25} established hardness of learning DNFs assuming search hardness of sparse learning parity with noise (sparse LPN). Hardness of learning DNFs also implies hardness for other classes, such as deterministic finite automata, polynomial threshold functions, and regular expressions~(cf. \citep{DSS16,DV21,ARSV25}). Moreover, since a DNF with $q(N)$ terms can be expressed as the complement of an intersection of $q(N)$ halfspaces, the above results imply hardness for intersections of halfspaces over $\bits^N$. A more direct hardness result for intersections of a polynomial number of halfspaces was shown by \citet{KS09}, assuming polynomial-time hardness of LWE. Finally, \citet{Tiegel24} proved hardness for intersections of $\omega(\log\log N)$ halfspaces over $\mathbb R^N$, assuming subexponential-time quantum hardness of either GapSVP or SIVP with polynomial approximation factors. We note that none of the above prior results applies to bounded-width DNFs or intersections of sparse halfspaces, which we study here, except for the junta hardness result of \citet{ABW10} discussed earlier.


We also note that our LWE assumptions follow from subexponential-time quantum hardness assumptions for worst-case lattice problems (see Section~\ref{sec:assumptions}). To our knowledge, even without a width bound, we obtain the first hardness results for improperly learning polynomial-size DNFs based on 
LWE or 
worst-case lattice hardness.

\paragraph{Implications for bounded-degree polynomial thresholds and \texorpdfstring{$\mathbb F_2$}{F2} polynomials.}
Under either LWE assumption, our junta lower bounds also apply to bounded-degree polynomial threshold functions over $\bits^N$ and $\mathbb F_2$ polynomials, with a polynomial number of monomials (see Section~\ref{sec:polynomial-results}).

\paragraph{Organization.}
Section~\ref{sec:preliminaries} defines the LWE assumptions and the PAC learning model. Section~\ref{sec:results} states the results, and Section~\ref{sec:proof-overview} discusses the proof ideas. Section~\ref{sec:discussion} presents some open questions.
Appendices~\ref{sec:proofs} and~\ref{sec:scope} contain the complete proofs of the junta theorems and their corollaries.

\section{Preliminaries}
\label{sec:preliminaries}

\paragraph{Notation.}
For a positive integer $m$, write $[m]=\{1,\ldots,m\}$. For $S\subseteq[N]$, $x_S$ denotes the restriction of $x\in\bits^N$ to $S$. All logarithms are base two.
We write $\Z_q$ for the integers modulo $q$, identified with $\{0,\ldots,q-1\}$. The notation $\poly$ denotes an unspecified polynomial. We denote the indicator function by $\onefunc[\cdot]$. We write $x\leftarrow S$ for a uniform draw from a finite set $S$, and $x\leftarrow\mathcal{D}$ for a sample from a distribution $\mathcal{D}$. A function $f:\mathbb Z_{\ge1}\to\mathbb Z_{\ge0}$ is \emph{efficiently computable} if a fixed deterministic algorithm, given $N$, outputs $f(N)$ in time polynomial in $N$. Throughout, $n$ denotes the LWE dimension and $N$ the number of Boolean variables in our learning problems.

\subsection{LWE and our assumptions}
\label{sec:lwe-distributions}\label{sec:assumptions}

Decisional learning with errors (LWE) requires distinguishing pairs \mbox{$(a,\langle a,s\rangle+e\bmod q)$} from uniformly random pairs, where $a$ is a uniformly random vector, $s$ is a common secret vector, and $e$ is a small random error.
LWE is a central and widely studied foundation of modern cryptography: standard hardness assumptions support public-key encryption~\citep{Reg09}, fully homomorphic encryption, and attribute-based encryption~\citep{Pei16}.
Our results use two quantitative LWE hardness assumptions. Below, we formally specify the distributions, state the assumptions and compare them with those used in prior work.

\paragraph{Parameters and distributions.}
For an integer $n$, let
\begin{equation}\label{eq:parameters}
 q=n^2,\qquad
 \ell=\lceil\log q\rceil,\qquad
 M=(n+1)\ell+2n~.
\end{equation}
Fix a sufficiently large integer $C\ge2$, independent of $n$, and let $\chi_n$ be the integer discrete Gaussian defined by
\begin{equation*}
 \chi_n(z)=\frac{\exp(-\pi z^2/(C^2 n))}{Z_n},\qquad
 Z_n=\sum_{a\in\Z}\exp(-\pi a^2/(C^2 n))~.
\end{equation*}
Define the LWE and uniform distributions on matrix-vector pairs by
\begin{align}
 (A,As+e\bmod q)&\sim\cL_n\colon
 &A&\leftarrow\Z_q^{M\times n},\quad
 s\leftarrow\Z_q^n,\quad e\leftarrow\chi_n^M~,
 \label{eq:lwe}\\
 (A,b)&\sim\cU_n\colon
 &A&\leftarrow\Z_q^{M\times n},\quad
 b\leftarrow\Z_q^M~.
 \label{eq:uniform}
\end{align}
Note that in the LWE distribution, the $M$ samples share one secret $s$.

\paragraph{Two quantitative hardness assumptions.}
In this work, we consider two commonly used assumptions that differ in their running-time requirements:

\begin{assumption}[$2^{o(n)}$-time LWE]
\label{ass:no-subexp}
There is no probabilistic algorithm $\mathcal B$ with worst-case running time $2^{o(n)}$ such that
\begin{equation}\label{eq:lwe-advantage}
 \left|
 \Pr_{(A,b)\sim\cL_n}[\mathcal B(A,b)=1]
 -\Pr_{(A,b)\sim\cU_n}[\mathcal B(A,b)=1]
 \right|>\frac1{10}
\end{equation}
for every sufficiently large $n$. Here $\cL_n,\cU_n$ are from~\eqref{eq:lwe}--(\ref{eq:uniform}).
\end{assumption}

\begin{assumption}[$2^{O(n^\beta)}$-time LWE]
\label{ass:lwe}
Fix a rational constant $\beta\in(1/2,1)$. There is no probabilistic algorithm $\mathcal B$ whose worst-case running time is $2^{O(n^\beta)}$ and for which~\eqref{eq:lwe-advantage} holds for every sufficiently large $n$.
\end{assumption}

In both statements, running time is worst-case over inputs and coins, and the probabilities include the algorithm's internal coins. Note that Assumption~\ref{ass:no-subexp} implies Assumption~\ref{ass:lwe} for every fixed rational $\beta\in(1/2,1)$. We emphasize that Theorem~\ref{thm:beta-main} and its corollaries require Assumption~\ref{ass:lwe} only for one fixed $\beta$, such as $\beta=0.51$, rather than for every fixed $\beta\in(1/2,1)$.

\paragraph{Relation to assumptions from prior learning-hardness results.}
Subexponential LWE assumptions (or lattice hardness assumptions that imply them) have been used to prove hardness of learning Gaussian mixtures~\citep{GVV22}, halfspaces with Massart noise~\citep{DKMR22}, and intersections of halfspaces~\citep{Tiegel24}, as well as agnostic learning of halfspaces and ReLU regression under Gaussian marginals~\citep{DKR23}, and learning from chain-of-thought data from multiple sources~\citep{JMSTV26}.


Assumption~\ref{ass:lwe} is implied by each of the assumptions in \citet[Assumption 2.4]{DKMR22} and \citet[Assumption 2.3]{DKR23}.
Their assumptions rule out distinguishers that use subexponentially many LWE samples and achieve inverse-subexponential advantage, while our assumption only rules out distinguishers that use $M=\Theta(n\log n)$ samples and achieve advantage greater than $1/10$.
Moreover, their assumptions imply Assumption~\ref{ass:lwe} for every fixed rational $\beta\in(1/2,1)$.
\citet[Theorem~2 and Figure~1]{GVV22} assume subexponential LWE hardness, even with subexponentially many LWE samples and inverse-subexponential advantage, but they allow a constant $\beta \in (0,1)$ while we require $\beta \in (1/2,1)$. 
The assumption of subexponential-time quantum hardness of GapSVP or SIVP with polynomial approximation factors in \citet[Assumption 4]{Tiegel24} implies Assumption~\ref{ass:no-subexp}, and hence Assumption~\ref{ass:lwe} for every fixed rational $\beta\in(1/2,1)$, by the quantum worst-case-to-average-case reduction for LWE~\citep[Theorem 4.2.4]{Pei16}.
The same assumption also appears in \citet[Hardness Assumption~1]{JMSTV26}.

Assumption~\ref{ass:no-subexp} specializes the exponential LWE assumption in \citet[Assumption LWE A1]{BBTV25} to modulus $q=n^2$, error distribution $\chi_n$, $M=\Theta(n\log n)$ samples, and advantage greater than $1/10$.
Finally, we note that our assumptions require stronger running-time hardness than the polynomial-time LWE hardness assumed by \citet{KS09}.

\subsection{PAC learning}
\label{sec:model}
We consider improper distribution-free realizable PAC learning from random examples.
A \emph{hypothesis class} is a sequence $\mathcal{H}=(\mathcal{H}_N)_{N\ge1}$, where each $\mathcal{H}_N$ is a collection of functions from a domain $\mathcal{X}_N$ to a label space $\mathcal{Y}$. For functions $f,h:\mathcal{X}_N\to\mathcal{Y}$ and a distribution $\mathcal{D}$ on $\mathcal{X}_N$, let
\[
 \err_{\mathcal{D},f}(h)=\Pr_{x\sim\mathcal{D}}[h(x)\ne f(x)]~.
\]
An \emph{example oracle} returns an independent pair $(x,f(x))$, with $x\sim\mathcal{D}$, on each request.

A randomized algorithm 
\emph{PAC learns} $\mathcal{H}$ if, for every $f\in\mathcal{H}_N$ and every distribution $\mathcal{D}$ on $\mathcal{X}_N$, given $\varepsilon,\delta\in(0,1)$ and access to an example oracle, it outputs a description of a hypothesis $h:\mathcal{X}_N\to\mathcal{Y}$ such that
\[
 \Pr\!\left[\err_{\mathcal{D},f}(h)\le\varepsilon\right]\ge1-\delta~,
\]
where the probability is over the examples and the learner's random choices.
We allow \emph{improper} learning, that is, the returned hypothesis $h$ need not belong to $\mathcal{H}_N$.
The algorithm's output is a finite binary string interpreted by a fixed deterministic evaluator $\Eval$, with $h(x)=\Eval(h,x)$. Note that we use the same notation for the description and the function it represents.
In this work, $\mathcal{X}_N=\bits^N$ and $\mathcal{Y}=\bits$.

A class is \emph{learnable in time $T(N,1/\varepsilon,1/\delta)$} if such a learner exists for which both the time to produce $h$ and the time to evaluate $h$ on one input are at most $T(N,1/\varepsilon,1/\delta)$, worst-case over all (possibly non-realizable) oracle responses and internal coins.
The class is \emph{polynomial-time learnable} if $T$ is polynomial in $N,1/\varepsilon,1/\delta$.

In our lower-bound statements and comparisons with upper bounds, we fix \mbox{$\varepsilon=\delta=1/10$} for convenience and denote the learning time by $T(N)$.
By a standard boosting argument~\citep{Sch90}, all our hardness results also hold for weak learning with error at most $1/2-1/\gamma(N)$, for any $\gamma(N)=\poly(N)$. The polynomial overhead introduced by boosting preserves our quantitative running-time bounds.

A Boolean function $f:\bits^N\to\bits$ is a $\kappa$-junta if there are $S\subseteq[N]$ with $|S|\le\kappa$, and $g:\bits^{|S|}\to\bits$ such that $f(x)=g(x_S)$ for every $x$.
A Boolean decision tree is a rooted full binary tree whose internal nodes query individual input coordinates and whose leaves carry Boolean labels.
On an input, computation follows the branches specified by the queried bits and outputs the label of the reached leaf.

\section{Results}
\label{sec:results}

In this section, we state the junta lower bounds and their consequences for decision trees, bounded-width DNFs, intersections of sparse halfspaces, and bounded-degree polynomial thresholds and $\mathbb F_2$ polynomials.

\subsection{Juntas under the \texorpdfstring{$2^{o(n)}$}{2\string^o(n)}-time LWE assumption}
\label{sec:no-subexp}

We first state our junta lower bounds under Assumption~\ref{ass:no-subexp}.

\begin{theorem}
\label{thm:main}\label{thm:no-subexp}
Under Assumption~\ref{ass:no-subexp}, the following hold.
\begin{enumerate}[label=\textnormal{(\roman*)}]
\item Suppose $\kappa(N)=\omega\bigl(\log\log N\,\log\log\log N\bigr)$ and is efficiently computable.
      Then $\kappa(N)$-juntas on $N$ variables are not polynomial-time improperly PAC learnable.
\item Suppose $\kappa(N)$ satisfies the conditions in part~\textnormal{(i)}, $\kappa(N)=O(\log N)$, and there is a constant $C_\kappa\ge1$ such that $\kappa(N+1)\le C_\kappa\kappa(N)$ for all sufficiently large $N$.
      Then $\kappa(N)$-juntas on $N$ variables are not learnable in time
      \begin{equation}\label{eq:runtime-no-subexp}
       T(N)=N^{o\!\left(\kappa(N)/(\log \kappa(N)\,\log\log N)\right)}~.
      \end{equation}
\end{enumerate}
\end{theorem}

The requirement $\kappa(N+1)\le C_\kappa\kappa(N)$ in part~\textnormal{(ii)} is a mild technical condition used in our reduction to rule out certain pathological behaviors of $\kappa$.\footnote{As an alternative to this condition, it suffices to require $\kappa(N)=\lfloor f(N)\rfloor$ for all sufficiently large $N$, where $f$ is an eventually nondecreasing, concave real-valued function. The same alternative applies to Theorem~\ref{thm:beta-main}\textnormal{(ii)}.}

\paragraph{Comparison with the best known upper bound.}
Distribution-free PAC learning of $\kappa$-juntas from random examples admits an $N^{O(\kappa)}$-time algorithm based on brute-force enumeration of candidate supports (cf.~\citep{Ser25}).\footnote{For each support, one checks whether the sample labels are consistent with a function of those coordinates and, if so, returns a truth table extending the observed labels.}
Theorem~\ref{thm:main}\textnormal{(ii)} excludes $N^{o(\kappa/(\log \kappa\,\log\log N))}$ time, and thus, the exponents differ by a factor $\log \kappa\,\log\log N$.
For $\kappa=\Theta((\log N)^\alpha)$, where $\alpha\in(0,1]$ is a fixed rational, the exclusion becomes $N^{o(\kappa/(\log\log N)^2)}$, within a factor $(\log\log N)^2$ in the exponent of the upper bound.

\subsection{Juntas under the \texorpdfstring{$2^{O(n^\beta)}$}{2\string^O(n\string^beta)}-time LWE assumption}
Next, we state the junta lower bounds under our weaker assumption.

\begin{theorem}
\label{thm:beta-main}
Under Assumption~\ref{ass:lwe} for some fixed rational $\beta\in(1/2,1)$, the following hold.
\begin{enumerate}[label=\textnormal{(\roman*)}]
\item Suppose $\kappa(N)=\Omega\bigl((\log N)^{1/\beta-1}\,(\log\log N)^2\bigr)$.
      Then $\kappa(N)$-juntas on $N$ variables are not polynomial-time improperly PAC learnable.
\item Suppose $\kappa(N)$ satisfies the condition in part~\textnormal{(i)} and is efficiently computable. Suppose also that $\kappa(N)=O(\log N)$ and there is a constant $C_\kappa\ge1$ such that $\kappa(N+1)\le C_\kappa\kappa(N)$ for all sufficiently large $N$.
      Then $\kappa(N)$-juntas on $N$ variables are not learnable in time
      \begin{equation}\label{eq:runtime-beta}
       T(N)=2^{O\!\left(\left[
        \frac{\kappa(N)\log N}{\log \kappa(N)\,\log\log N}
       \right]^\beta\right)}~.
      \end{equation}
\end{enumerate}
\end{theorem}

For example, the threshold in part~\textnormal{(i)} is $\sqrt{\log N}\,(\log\log N)^2$ when $\beta=2/3$, and $(\log N)^{1/3}\,(\log\log N)^2$ when $\beta=3/4$.
If Assumption~\ref{ass:lwe} holds for \emph{every} fixed rational $\beta\in(1/2,1)$ (as assumed in \citep{DKMR22} and \citep{DKR23}), then part~\textnormal{(i)} rules out polynomial-time learning of $(\log N)^\alpha$-juntas for \emph{every fixed constant $\alpha>0$}.

\subsection{Decision trees}
\label{sec:explicit-runtime}
Every $\kappa$-junta can be expressed by a decision tree with at most $2^\kappa$ leaves.
Hence, applying part~\textnormal{(i)} of Theorems~\ref{thm:main} and~\ref{thm:beta-main} gives the following corollary.
Note that both parts of the corollary already give hardness for trees with a subpolynomial number of leaves: they allow $\kappa(N)=o(\log N)$, and then $2^{\kappa(N)}=N^{o(1)}$.

\begin{corollary}[Polynomial-time hardness for decision trees]
\label{cor:trees}
\leavevmode
\begin{enumerate}[beginpenalty=10000,label=\textnormal{(\roman*)}]
\item Under Assumption~\ref{ass:no-subexp}, if $\kappa(N)=\omega(\log\log N\,\log\log\log N)$ and is efficiently computable, then decision trees with at most $2^{\kappa(N)}$ leaves are not polynomial-time improperly PAC learnable.
\item Under Assumption~\ref{ass:lwe} for some fixed rational $\beta\in(1/2,1)$, if $\kappa(N)=\Omega((\log N)^{1/\beta-1}\,(\log\log N)^2)$, then decision trees with at most $2^{\kappa(N)}$ leaves are not polynomial-time improperly PAC learnable.
\end{enumerate}
In particular, each assumption rules out polynomial-time improper PAC learning of decision trees with at most $N$ leaves. Thus, it suffices that Assumption~\ref{ass:lwe} holds for a single $\beta$ (e.g., $\beta=0.51$).
\end{corollary}

The quantitative lower bound in Theorem~\ref{thm:main}\textnormal{(ii)} also applies to decision trees with at most $2^{\kappa(N)}$ leaves:

\begin{corollary}[Quantitative hardness for decision trees]
\label{cor:tree-runtime}
Under Assumption~\ref{ass:no-subexp}, if $\kappa(N)$ satisfies the conditions of Theorem~\ref{thm:main}\textnormal{(ii)}, then decision trees with at most $2^{\kappa(N)}$ leaves are not learnable in time $T(N)=N^{o\!\left(\kappa(N)/(\log \kappa(N)\,\log\log N)\right)}$.
\end{corollary}

For the analogous quantitative corollary under Assumption~\ref{ass:lwe}, see Corollary~\ref{cor:tree-runtime-beta}.

\paragraph{Comparison with the best known upper bound.}
The best known upper bound for distribution-free learning of $s$-leaf Boolean decision trees from random examples is $N^{O(\log s)}$, achieved by the algorithm of \citet{EH89}. Writing $s=2^{\kappa(N)}$, Corollary~\ref{cor:tree-runtime} excludes $N^{o(\log s/(\log\log s\,\log\log N))}$ time.
Thus, the exponents in this upper bound and our excluded runtime differ by a factor $\log\log s\,\log\log N$.
For $s=N$, the upper bound is $N^{O(\log N)}$, whereas our result excludes $N^{o(\log N/(\log\log N)^2)}$ time.

\subsection{Bounded-width DNFs and intersections of sparse halfspaces}
\label{sec:dnf}

\paragraph{Bounded-width DNFs.}
A DNF on $\bits^N$ is a disjunction of terms, each of which is a conjunction of literals. Its width is the maximum number of literals in a term.
A $\kappa$-DNF is a DNF of width at most $\kappa$.
Every $\kappa$-junta can be expressed by a $\kappa$-DNF with at most $2^\kappa$ terms: include one term specifying each accepting assignment to its relevant coordinates.
Hence, applying Theorem~\ref{thm:main} gives the following corollary.

\begin{corollary}[Hardness for bounded-width DNFs]\label{cor:dnf}
Under Assumption~\ref{ass:no-subexp}, if $\kappa(N)=\omega(\log\log N\,\allowbreak\log\log\log N)$ and is efficiently computable, then $\kappa(N)$-DNFs on $\bits^N$ with at most $2^{\kappa(N)}$ terms are not polynomial-time improperly PAC learnable.
If, in addition, $\kappa(N)$ satisfies the conditions of Theorem~\ref{thm:main}\textnormal{(ii)}, then they are not learnable in time $T(N)=N^{o\!\left(\kappa(N)/(\log \kappa(N)\,\log\log N)\right)}$.
\end{corollary}

Note that for $\kappa(N)=O(\log N)$, these DNFs have at most polynomially many terms.
For the analogous result under Assumption~\ref{ass:lwe}, see Corollary~\ref{cor:dnf-beta}.

\paragraph{Intersections of sparse halfspaces.}
A $\kappa$-sparse (affine) halfspace on $\bits^N$ is a Boolean function $x\mapsto\onefunc[\langle v,x\rangle\ge\theta]$, where $v\in\mathbb R^N$, $\theta\in\mathbb R$, and $|\{i:v_i\ne0\}|\le\kappa$.
A simple reduction transfers the junta lower bounds to intersections of halfspaces over the hypercube with weights in $\{0,1\}$, threshold one, and a bound on the union of their supports. Appendix~\ref{app:sparse-halfspaces} proves the following corollary and the analogous result under Assumption~\ref{ass:lwe}.

\begin{corollary}[Hardness for intersections of sparse halfspaces]
\label{cor:sparse-halfspaces}
Under Assumption~\ref{ass:no-subexp}, if $\kappa(N)=\omega(\log\log N\,\log\log\log N)$ and is efficiently computable, then intersections of at most $2^{\kappa(N)}$ halfspaces on $\bits^N$, each of which is $\kappa(N)$-sparse, are not polynomial-time improperly PAC learnable.
If, in addition, $\kappa(N)$ satisfies the conditions of Theorem~\ref{thm:main}\textnormal{(ii)}, then they are not learnable in time $T(N)=N^{o\!\left(\kappa(N)/(\log \kappa(N)\,\log\log N)\right)}$.
Both bounds hold even when every weight lies in $\{0,1\}$, every threshold equals $1$, and the union of the halfspaces' supports has size at most $\kappa(N)$.
\end{corollary}

\paragraph{Comparison with the best known upper bounds.}
An $N^{O(\kappa)}$ upper bound for learning $\kappa$-DNFs was already shown by \citet{Val84}.
It also implies the same bound for intersections of $\kappa$-sparse halfspaces on $\bits^N$, since they admit $\kappa$-CNF representations.\footnote{For each halfspace and each falsifying assignment $a$ to its relevant coordinates, form the clause with literal $x_i$ if $a_i=0$ and $\neg x_i$ if $a_i=1$. The conjunction of all these clauses represents the intersection and each clause has width at most $\kappa$.}
Thus, the exponent differs from that in our excluded runtimes under Assumption~\ref{ass:no-subexp} by a factor $\log \kappa\,\log\log N$.\footnote{\citet{KS01} give a running time of $N^{O(\sqrt{\kappa}\,\log s)}$ for learning $\kappa$-DNFs with $s$ terms. \citet{KOS04} give the same bound for intersections of $s$ halfspaces, each of integer weight at most $\kappa$. When $s=2^\kappa$, this bound is looser than $N^{O(\kappa)}$.}

\subsection{Bounded-degree polynomial thresholds and \texorpdfstring{$\mathbb F_2$}{F2} polynomials}
\label{sec:polynomial-results}
Every $\kappa$-junta on $\bits^N$ can be represented both as a real polynomial threshold function and as a polynomial over $\mathbb F_2$.
In each representation, the polynomial has degree at most $\kappa$ and at most $2^\kappa$ monomials. Thus, under either LWE assumption, the same polynomial-time hardness thresholds and running-time lower bounds apply to learning these classes. Note that when $\kappa = O(\log N)$, at most $N^{O(1)}$ monomials suffice.
Appendix~\ref{app:sparse-polynomials} gives the formal statement.

\section{Proof overview}
\label{sec:proof-overview}

In this section we provide an overview of the main proof ideas (see Appendix~\ref{sec:proofs} for the full proofs).
Our starting point is the approach of \citet{KS09}: learning the decryption function of Regev's encryption scheme \citep{Reg09} would break its security and allow for an LWE distinguisher.
Encryptions of randomly chosen bits give labeled examples whose target is decryption with a fixed secret key. To use a junta learner, we must make decryption depend on few coordinates without making examples too expensive to generate, and the reduction must also remain valid when the learner uses subexponentially many examples. Steps~1--4 below build this reduction, and Steps~5--6 
derive the hardness results and running-time bounds.

\paragraph{1. Generating labeled examples by encryption.}
Let $n$ be the LWE dimension, $q=n^2$ the modulus, and $M=\Theta(n\log n)$ the number of LWE samples (as in~\eqref{eq:parameters}). Choose a uniform secret $s\in\Z_q^n$, an independent uniform matrix $A\in\Z_q^{M\times n}$, and an independent error vector $e$ from the LWE error distribution in Section~\ref{sec:lwe-distributions}.
In Regev's encryption, the public key is the LWE instance $(A,b)$, where $b=As+e\pmod q$, and $s$ remains secret.
To encrypt a bit $\mu$, choose $\rho\in\bits^M$ uniformly at random and output the ciphertext $(u,w)$ given by
\[
 u=A^\top\rho\pmod q,\qquad
 w=b^\top\rho+\mu c_q\pmod q~,
\]
where $c_q=\lfloor q/2\rfloor$.
These operations use only the public key. To decrypt, the secret-key holder subtracts $\langle u,s\rangle$ from $w$. Substitution of $b=As+e$ cancels the contribution of $A$ and gives
\[
 w-\langle u,s\rangle=e^\top\rho+\mu c_q\pmod q~.
\]
Treating residues modulo $q$ as positions on a circle, return $0$ if this residue is closer to $0$, and $1$ if it is closer to $c_q$. Thus the bit is recovered whenever the noise $e^\top\rho$ is sufficiently small.

With probability $1-2^{-\Omega(n\log n)}$ over key generation, $\|e\|_2^2\le4C^2nM$. On this \emph{good-key event},
\[
 |e^\top\rho|\le\|e\|_1\le\sqrt M\|e\|_2
 \le2CM\sqrt n=O(n^{3/2}\log n)<q/8
\]
for \emph{every} $\rho$ once $n$ is sufficiently large.
Thus, on the good-key event, every encryption decrypts correctly, regardless of how many encryptions we generate, and without requiring a union bound over the generated encryptions.
Drawing $\mu\in\bits$ and $\rho\in\bits^M$ independently and uniformly gives labeled examples $\bigl((u,w),\mu\bigr)$, which are realizable by the decryption function.

\paragraph{2. Expanding the input lets decryption read fewer entries.}
Decryption as written above may depend on all coordinates of $u$. We replace the ciphertext by tables of possible block contributions to $\langle u,s\rangle$. Choose a block length $1\le t\le\lfloor n/2\rfloor$ and partition the coordinates of $u$ into $J=\lceil n/t\rceil$ blocks, padding the last block if necessary. For each block $j$, form
\[
 \bigl(\langle u^{(j)},a\rangle\bmod q\bigr)_{a\in\Z_q^t}~.
\]
This table is computed from $u$ alone and enumerates \emph{possible secret blocks}. The actual secret block $s^{(j)}$ determines which entry decryption reads. The $J$ selected entries sum to the secret inner product modulo $q$, and together with $w$, they suffice to decrypt.
Note that their values vary across examples, but their positions depend only on the fixed secret key.

Binary encoding of the tables and $w$ gives a Boolean feature map of dimension \mbox{$(1+Jq^t)\lceil\log q\rceil$} where the decryption function reads $J+1$ encoded values, each represented by $\lceil\log q\rceil=\Theta(\log n)$ bits. Increasing $t$ reduces the number of selected values but increases the number $q^t$ of entries in each table.
Our actual feature map rounds $w$ and every table entry to save bits, as discussed below.

\paragraph{3. Rounding uses the decryption margin to save bits.}
On the good-key event, the decryption residue is within $q/8$ of the center corresponding to the encrypted bit. Since the two centers are nearly $q/2$ apart, small rounding errors do not change which center is nearer.
This margin allows us to round $w$ and every table entry to $r=\Theta(\log J)$ bits while preserving correct decryption.
The resulting decryption function depends on only
\[
 K=(J+1)r=\Theta\bigl(J\log J\bigr)
\]
coordinates, so the generated examples remain realizable by a $K$-junta.
Thus, rounding reduces the number of relevant bits from $\Theta(J\log n)$ to $\Theta(J\log J)$.

\paragraph{4. A learner gives an LWE distinguisher.}
We use the statistical hiding argument of \citet{Reg09} to turn learning into LWE distinguishing. 
Note that we can use a junta learner to break the encryption scheme as follows.
With a fixed public key, encrypt independent fair bits, apply the feature map, and give the resulting labeled examples to the learner. On the good-key event, the labels are realizable by a $K$-junta. To predict the bit encrypted in a fresh ciphertext, apply the feature map and evaluate the learned hypothesis. Thus, a junta learner allows us to obtain a plaintext predictor.

The distinguisher uses its input $(A,b)$ as the public key. It encrypts a fresh fair bit $\mu$ and accepts if the above plaintext predictor recovers $\mu$. On an LWE input, the learning guarantee gives an acceptance probability exceeding $1/2$ by a constant.
On a uniform input, jointly with the public key and $\mu$, the fresh ciphertext is statistically close to an independent uniform vector. In this ideal experiment where the ciphertext is independent of $\mu$, the predictor succeeds with probability exactly $1/2$. The distinguisher's acceptance probability on a uniform input is therefore at most $1/2$ plus the statistical error.

Generating the training examples from the public key and running the learner cannot increase this statistical distance.
Thus the statistical error is independent of the number of generated examples.
Together with Step~1's correctness guarantee for every encryption, this allows the reduction to use subexponentially many training examples.

\paragraph{5. Choosing the learning dimension and proving polynomial-time hardness.}
Longer blocks let decryption read fewer entries but produce larger tables for each example. Choosing the block length so that the feature map has at most $N$ coordinates gives
\begin{equation}\label{eq:overview-dimension-tradeoff}
 J=\Theta\!\left(\frac{n\log n}{\log N}\right),\qquad
 K=\Theta(J\log J)~.
\end{equation}
To prove polynomial-time hardness under Assumption~\ref{ass:no-subexp}, assume that a learner for $\kappa(N)$-juntas runs in polynomial time $T(N)$, where $\kappa(N)=\omega(\log\log N\,\log\log\log N)$ and is efficiently computable.
The reduction must find a dimension $N$ with $K(n,N)\le\kappa(N)$, so that the target from Step~3 lies in the learner's class, and with $N=2^{o(n)}$, so that the distinguisher from Step~4 runs in time $2^{o(n)}$.
Such a dimension can be found by search: we examine integer dimensions in increasing order, computing $\kappa(N)$ in time polynomial in $N$ for each candidate.
We stop at the first dimension satisfying $K(n,N)\le\kappa(N)$ and denote it by $N(n)$. For every fixed $c\in(0,1)$, a dimension $N$ of order $2^{cn}$ is eventually feasible as the construction uses $K=O(\log n\,\log\log n)$ relevant bits by~\eqref{eq:overview-dimension-tradeoff}, and $\kappa(N)=\omega(\log\log N\,\log\log\log N) =\omega(\log n\,\log\log n)$ grows faster. Since this holds for every fixed $c$, the first feasible dimension satisfies \mbox{$N(n)=2^{o(n)}$}.
The distinguisher takes $\poly(n,N)$ time to find $N$ and at most $N\cdot T(N)\poly(n)$ time for the remaining computation.
Since $T(N)$ is polynomial, the distinguisher's total running time is $2^{o(n)}$, contradicting Assumption~\ref{ass:no-subexp}.

For polynomial-time hardness under Assumption~\ref{ass:lwe}, assume that a learner for $\kappa(N)$-juntas runs in polynomial time, where $\kappa(N)=\Omega((\log N)^{1/\beta-1}\,(\log\log N)^2)$.
Here, we choose $N(n)$ explicitly on the scale $2^{cn^\beta}$, with a fixed integer $c$.
Taking $c$ large enough can make the constructed junta size $K$ fall below any fixed positive multiple of $(\log N)^{1/\beta-1}\,(\log\log N)^2$ for all sufficiently large $n$, and we can therefore ensure $K\le\kappa(N)$. Running a polynomial-time learner in dimension $N(n)$ still costs $2^{O(n^\beta)}$, yielding a distinguisher that contradicts Assumption~\ref{ass:lwe}.
Note that this choice of $N$ never evaluates $\kappa$, so the result needs no computability assumption on it.

\paragraph{6. Proving quantitative time lower bounds.}
For quantitative lower bounds, we again use the first dimension $N=N(n)$ found by the search of Step~5. The assumptions $\kappa(N)=O(\log N)$ and $\kappa(N+1)\le C_\kappa\kappa(N)$ imply that $K(n,N)=\Theta(\kappa(N))$. Indeed, since $N$ is the first feasible dimension, the constructed junta is too large at dimension $N-1$ but fits at dimension $N$. Between these two dimensions, $\kappa$ cannot increase by more than a constant factor, and $K$ cannot decrease by more than a constant factor.
Combining this comparison with~\eqref{eq:overview-dimension-tradeoff} gives
\[
 n=\Theta\!\left(
 \frac{\kappa(N)\log N}{\log\kappa(N)\,\log\log N}
 \right)~.
\]
This expresses the LWE dimension in terms of the learner's parameters.
A learner with either of the running-time bounds ruled out by Theorems~\ref{thm:main}\textnormal{(ii)} and~\ref{thm:beta-main}\textnormal{(ii)} would therefore give an LWE distinguisher running in $2^{o(n)}$ or $2^{O(n^\beta)}$ time, respectively, contradicting the corresponding LWE assumption.

\section{Discussion}
\label{sec:discussion}

Our results give LWE-based lower bounds for learning sublogarithmic juntas and subpolynomial-size decision trees, as well as implications for additional classes.

Several questions on the computational complexity of improperly learning juntas and decision trees are left open: Can hardness under Assumption~\ref{ass:no-subexp} be extended to $O(\log\log N)$-juntas or decision trees with $(\log N)^{O(1)}$ leaves?
Can hardness be shown for $\kappa$-juntas already for any $\kappa=\omega(1)$? Can we narrow the gap between the lower bound of Theorem~\ref{thm:main}\textnormal{(ii)} and the $N^{O(\kappa)}$ upper bound?
Finally, can analogous hardness results be established for learning juntas or decision trees with examples drawn uniformly from $\bits^N$?

\subsection*{AI Use Statement}
AI tools, primarily GPT-6 Astra, were used to assist in developing the theoretical results, identifying relevant literature, and drafting and editing the manuscript. The author reviewed, verified, and revised all AI-generated material and is responsible for the content of the paper.

\subsection*{Acknowledgements}
This research is supported by the Israel Science Foundation (grant No. 2574/25), by a research grant from Mortimer Zuckerman (the Zuckerman STEM Leadership Program), and by research grants from the Center for New Scientists at the Weizmann Institute of Science, and the Shimon and Golde Picker -- Weizmann Annual Grant.

\printbibliography[heading=bibintoc]

\clearpage
{\small\tableofcontents}
\clearpage
\appendix

\section{Proofs}
\label{sec:proofs}\label{app:construction}

We prove Theorems~\ref{thm:main} and~\ref{thm:beta-main} by encoding ciphertexts in LWE dimension $n$ as Boolean examples in dimension $N$.
Sections~\ref{sec:correctness}--\ref{sec:encoding} 
construct a junta that correctly decrypts every ciphertext generated under a good key.
Sections~\ref{sec:encryption-security}--\ref{sec:reduction} turn a learner into an LWE distinguisher, and Sections~\ref{sec:proof-beta-poly}--\ref{sec:proof-explicit-time} choose $N$ and conclude the proofs.

Table~\ref{tab:notation} summarizes the parameters:

\begin{table}[htbp]
\centering
\small
\renewcommand{\arraystretch}{1.08}
\begin{tabular}{@{}ll@{}}
\toprule
Symbol & Meaning \\
\midrule
$n$ & LWE dimension \\
$N$ & Boolean input dimension, chosen as $N(n)$ \\
$q,\ell$ & LWE modulus and bits per residue: $q=n^2$, $\ell=\lceil\log q\rceil$ \\
$M$ & Number of LWE samples: $M=(n+1)\ell+2n$ \\
$t,J$ & Block length and number of blocks: $J=\lceil n/t\rceil$ \\
$r,B$ & Rounding precision: $r=\lceil\log (64J)\rceil$ bits, $B=2^r$ levels \\
$N_0$ & Feature dimension before padding: $N_0=r(1+Jq^t)$ \\
$K$ & Number of selected bits (junta-size bound): $K=(J+1)r$ \\
$\kappa(N)$ & Learner's bound on the number of relevant bits \\
$\beta$ & Fixed rational exponent in the LWE time bound, $1/2<\beta<1$ \\
\bottomrule
\end{tabular}
\caption{Parameters used in the proofs.}
\label{tab:notation}
\end{table}

\FloatBarrier

\subsection{Encryption and simultaneous correctness}
\label{sec:correctness}\label{app:correctness}

For any fixed matrix-vector pair $(A,b)$ with the dimensions specified in Section~\ref{sec:lwe-distributions}, define
\begin{equation}\label{eq:enc}
 \Enc_{A,b}(\mu;\rho)=(u,w),\qquad
 u=A^\top\rho\pmod q,\quad
 w=b^\top\rho+\mu c_q\pmod q~,
\end{equation}
where $c_q=\lfloor q/2\rfloor$, $\rho\in\bits^M$, and $\mu\in\bits$. A labeled sample consists of this ciphertext and label $\mu$, with $\rho$ and $\mu$ independent and uniform. The procedure uses $(A,b)$ alone and is defined even when $(A,b)$ is uniform.

For $a,b\in\Z_q$, let
\[
 d_q(a,b)=\min_{j\in\Z}|a-b+jq|
\]
be circular distance. For each $s\in\Z_q^n$, define a decryption function by
\begin{equation*}
 \Dec_s(u,w)=
 \begin{cases}
 0,&d_q(w-\langle u,s\rangle,0)
       \le d_q(w-\langle u,s\rangle,c_q),\\
 1,&\text{otherwise}~.
 \end{cases}
\end{equation*}
Ties are resolved in favor of $0$, although no ties occur in the correctness argument.

Denote by
\[
 \mathcal G_n=\{e\in\Z^M:\|e\|_2^2\le4C^2nM\}
\]
the good event on the error vector of a generating triple $(A,s,e)$ defined in Section~\ref{sec:lwe-distributions}. Below we bound its failure probability, and show that on $\mathcal G_n$ the function $\Dec_s$ correctly decrypts every possible encryption.

\begin{lemma}[Simultaneous correctness and decryption margin]
\label{lem:correctness}\label{lem:tail}
Let $\gamma_n=e^{-(2\pi-\ln4)M}$. For $e\leftarrow\chi_n^M$,
\begin{equation}\label{eq:goodprob}
 \Pr[e\notin\mathcal G_n]\le\gamma_n=2^{-\Omega(n\log n)}~.
\end{equation}
Moreover, for all sufficiently large $n$, every generating triple $(A,s,e)$ with $e\in\mathcal G_n$ and $b=As+e\pmod q$ satisfies $|e^\top\rho|\le2CM\sqrt n<q/8$ for every $\rho\in\bits^M$.
Consequently,
\begin{equation*}
 \Dec_s(\Enc_{A,b}(\mu;\rho))=\mu
 \quad\text{for every }\rho\in\bits^M
 \text{ and every }\mu\in\bits~.
\end{equation*}
\end{lemma}
\begin{proof}\leavevmode
\paragraph{Probability of the good event.}
Let $\sigma=C\sqrt n$. We first bound the normalizing constant $Z_n$ from Section~\ref{sec:lwe-distributions} by comparing its defining sum with an integral.
The function $g(t)=\exp(-\pi t^2/\sigma^2)$ is decreasing for $t\ge0$, so, on each interval $[k,k+1]$ with integer $k\ge0$, $g(k+1)\le g(t)\le g(k)$.
Integrating over these intervals and summing gives
\[
 \sum_{k=1}^{\infty}g(k)
 \le\int_0^\infty g(t)\,dt
 =\frac\sigma2
 \le\sum_{k=0}^{\infty}g(k)
 =1+\sum_{k=1}^{\infty}g(k)~,
\]
where the first equality follows from the Gaussian integral and the second uses $g(0)=1$.
Since $g$ is even, we have $Z_n=1+2\sum_{k=1}^{\infty}g(k)$, and hence $\sigma-1\le Z_n\le\sigma+1$.

For any fixed $i\in[M]$, the definition of $\chi_n$ and the bound above applied at $n$ and $2n$ give
\[
 \begin{aligned}
 \mathbb E\exp\!\left(\frac{\pi e_i^2}{2\sigma^2}\right)
 &=\frac1{Z_n}\sum_{z\in\Z}
   \exp\!\left(-\frac{\pi z^2}{\sigma^2}\right)
   \exp\!\left(\frac{\pi z^2}{2\sigma^2}\right)\\
 &=\frac1{Z_n}\sum_{z\in\Z}
   \exp\!\left(-\frac{\pi z^2}{2C^2n}\right)
 =\frac{Z_{2n}}{Z_n}
 \le\frac{\sqrt2\sigma+1}{\sigma-1}\le4~,
 \end{aligned}
\]
where the last inequality uses $\sigma\ge2$.
Set $X=\exp(\pi\|e\|_2^2/(2\sigma^2))$.
Applying Markov's inequality and then using the independence of the coordinates of $e$, we obtain
\[
 \begin{aligned}
 \Pr[\|e\|_2^2>4\sigma^2M]
 &=\Pr[X>e^{2\pi M}]\le e^{-2\pi M}\,\E X\\
 &=e^{-2\pi M}\,\E\!\left[
   \prod_{i=1}^M\exp\!\left(\frac{\pi e_i^2}{2\sigma^2}\right)\right]
 =e^{-2\pi M}\prod_{i=1}^M
   \E\exp\!\left(\frac{\pi e_i^2}{2\sigma^2}\right)\\
 &\le4^M e^{-2\pi M}=\gamma_n~,
 \end{aligned}
\]
proving~\eqref{eq:goodprob}.

\paragraph{One margin bound for every encryption.}
Substituting $b=As+e\pmod q$ into~\eqref{eq:enc} gives
\begin{equation}\label{eq:cancel}
 w-\langle u,s\rangle=e^\top\rho+\mu c_q\pmod q~.
\end{equation}
On $\mathcal G_n$, every $\rho\in\bits^M$ satisfies
\[
 |e^\top\rho|\le\|e\|_1\le\sqrt M\|e\|_2
 \le2CM\sqrt n=O(n^{3/2}\log n)<q/8
\]
for all sufficiently large $n$, by~\eqref{eq:parameters}.
Write $\tau=|e^\top\rho|$. The circular distance from the residual in~\eqref{eq:cancel} to the correct center $\mu c_q$ is at most $\tau$. The two centers are at distance $c_q=\lfloor q/2\rfloor$, so the other center is at distance at least $c_q-\tau>\tau$.
The correct center is therefore uniquely closest for every $\rho,\mu$.
\end{proof}

\subsection{Encoding ciphertexts for junta learning}
\label{sec:encoding}\label{sec:construction}

Fix any integer block length $1\le t\le\lfloor n/2\rfloor$, and let
\begin{equation*}
 J=\lceil n/t\rceil\ge2~.
\end{equation*}
Pad $u,s\in\Z_q^n$ with zeros to length $Jt$ and partition them into $J$ consecutive blocks of length $t$. Then
\begin{equation}\label{eq:blocksum}
 \langle u,s\rangle
 =\sum_{j=1}^J\langle u^{(j)},s^{(j)}\rangle\pmod q~.
\end{equation}

\subsubsection{Block enumeration}
\label{sec:dictionary}

For each block $j$, list the inner products
\[
 \bigl(\langle u^{(j)},a\rangle\bmod q\bigr)_{a\in\Z_q^t}~.
\]
The secret block $s^{(j)}$ selects one entry of this table.
By~\eqref{eq:blocksum}, the $J$ selected entries determine $\langle u,s\rangle\bmod q$. The reduction can compute every table from $u$ without knowing the secret or which entries it selects.
The selected indices depend only on $s$, so the same selection serves every ciphertext.

Encoding the selected residues and the ciphertext component $w$ exactly uses $(J+1)\ell$ bits, with $\ell=\Theta(\log n)$. We instead round every table entry and $w$.
The decryption function then reads only $J+1$ rounded words. We will next show that we can keep the total rounding error below the decryption margin using $O(\log J)$ bits per residue instead of $\ell$. Thus rounding reduces the number of bits read by the decryption function from $(J+1)\ell$ to $O(J\log J)$.

\subsubsection{Rounded Boolean features}

We choose $B\ge64J$ to be a power of two as follows:
\begin{equation}\label{eq:precision}
 r=\left\lceil\log (64J)\right\rceil,
 \qquad B=2^r~.
\end{equation}
For a residue $a\in\Z_q$, we set
\[
 Q(a)=\left\lfloor\frac{Ba}{q}\right\rfloor
 \in\{0,\ldots,B-1\}~.
\]
Write $\bin_r(v)$ for the ordinary $r$-bit encoding of an integer $0\le v<B$. Define $\Phi_0(u,w)$ by concatenating
\[
 \bin_r(Q(w))
 \quad\text{and}\quad
 \bigl(\bin_r(Q(\langle u^{(j)},v\rangle\bmod q))\bigr)_{
       j\in[J],\,v\in\Z_q^t}~,
\]
in fixed lexicographic order. Its dimension is
\begin{equation}\label{eq:N0}
 N_0=r(1+Jq^t)~.
\end{equation}
For any integer $N\ge N_0$, let $\Phi:\Z_q^{n+1}\to\bits^N$ append zeros to $\Phi_0$. The map depends on $n,t,N$, but on neither the public key nor the secret.

Rounding might change decryption on arbitrary ciphertexts near a decision boundary. The following lemma shows that decryption remains correct on every ciphertext generated under a good key, as required for exact realizability.

Set
\begin{equation}\label{eq:junta-size}
 K=(J+1)r~.
\end{equation}

\begin{lemma}[Rounded encoded decryption]\label{lem:junta}
For all sufficiently large $n$, every integer $1\le t\le\lfloor n/2\rfloor$, every $N\ge N_0$, and every $s\in\Z_q^n$, there is a Boolean $K$-junta $f_s$ on $N$ coordinates with the following property. For every $A\in\Z_q^{M\times n}$ and $e\in\Z^M$ with $e\in\mathcal G_n$, setting $b=As+e\pmod q$ gives
\begin{equation}\label{eq:junta-dec}
 f_s\bigl(\Phi(\Enc_{A,b}(\mu;\rho))\bigr)=\mu
 \quad\text{for every }\rho\in\bits^M
 \text{ and every }\mu\in\bits~.
\end{equation}
The map $\Phi$ is computable in $N\poly(n)$ time, where the polynomial is independent of $t$ and $N$.
\end{lemma}
\begin{proof}\leavevmode
\paragraph{Defining $f_s$.}
On any Boolean input $x\in\bits^N$, read its first $r$-bit word and, for each block $j$, the word indexed by $(j,s^{(j)})$.
Decode these words as integers $w_x,a_{1,x},\ldots,a_{J,x}\in\{0,\ldots,B-1\}$ and let
\[
 z_x=w_x-\sum_{j=1}^J a_{j,x}\pmod B~.
\]
Define $f_s(x)$ by the nearest-center rule modulo $B$:
\[
 f_s(x)=
 \begin{cases}
 0,&d_B(z_x,0)\le d_B(z_x,B/2),\\
 1,&\text{otherwise}~.
 \end{cases}
\]
Here $d_B$ is the circular distance defined in Section~\ref{sec:correctness}, with modulus $B$ in place of $q$.
Note that the integer $B$ is even. This rule is defined for every Boolean input, including those whose words are mutually inconsistent as ciphertext features. It reads only $(J+1)r=K$ coordinates.

\paragraph{Preserving every encryption label.}
We now prove~\eqref{eq:junta-dec}. Fix a triple $(A,s,e)$ as in the statement and any encryption coins $\rho,\mu$.
For the resulting ciphertext $(u,w)$, write $a_j=\langle u^{(j)},s^{(j)}\rangle\bmod q$, and take the canonical integer representatives of $w,a_1,\ldots,a_J$.
For $a,b\in\mathbb R$, define their circular distance modulo $1$ by
\[
 d_\circ(a,b)=\min_{m\in\Z}|a-b+m|~.
\]
Dividing residues and centers by their modulus preserves the nearest-center rule. We therefore compare the exact and rounded decryption residues after normalizing each modulus to $1$. Set
\[
 z=\frac{Q(w)-\sum_jQ(a_j)}B,
 \qquad v=\frac{w-\sum_j a_j}{q}~.
\]
The definition of $Q$ gives $0\le a/q-Q(a)/B<1/B$ for every $a\in\Z_q$. Hence,
\[
 \begin{aligned}
 d_\circ(z,v)&\le |z-v|\\
 &\le \left|\frac{Q(w)}B-\frac wq\right|
   +\sum_{j=1}^J\left|\frac{Q(a_j)}B-\frac{a_j}q\right|\\
 &<\frac1B+\frac JB=\frac{J+1}{B}\le\frac3{128}<\frac1{32}~,
 \end{aligned}
\]
where we used $B\ge64J$ and $J\ge2$.
By~\eqref{eq:cancel} and~\eqref{eq:blocksum}, $v$ differs from $e^\top\rho/q+\mu c_q/q$ by an integer.
Since $|e^\top\rho|<q/8$ by Lemma~\ref{lem:correctness} and $|c_q/q-1/2|\le1/(2q)$, the triangle inequality gives
\begin{equation*}
 \begin{aligned}
 d_\circ(z,\mu/2)
 &\le d_\circ(z,v)+d_\circ(v,\mu c_q/q)
      +d_\circ(\mu c_q/q,\mu/2)\\
 &\le d_\circ(z,v)+\frac{|e^\top\rho|}{q}
      +\mu\left|\frac{c_q}{q}-\frac12\right|\\
 &<\frac1{32}+\frac18+\frac1{2q}<\frac14
 \end{aligned}
\end{equation*}
for all sufficiently large $n$. Since $d_\circ(0,1/2)=1/2$, the nearer center is uniquely $\mu/2$.
Equivalently, the nearer center for $z_x$ modulo $B$ is $\mu B/2$. Hence $f_s$ returns $\mu$.
Every bound holds simultaneously for all $\rho,\mu$ and every $1\le t\le\lfloor n/2\rfloor$.

\paragraph{Computing the feature map.}
Since $J\le n$, we have $r=\lceil\log(64J)\rceil=O(\log n)$.
Each table entry can be computed in $\poly(n)$ time: it requires an inner product of two vectors of length $t\le n$ modulo $q=n^2$, followed by rounding and binary encoding. There are $Jq^t\le N$ entries.
Computing these entries, rounding and encoding $w$, and padding the output to $N$ bits therefore takes $N\poly(n)$ bit operations, with a polynomial independent of $t$ and $N$.
\end{proof}

\subsubsection{The dimension--support tradeoff}

We now choose the block length from the learning dimension $N$.
Larger $N$ allows longer blocks, so the decryption reads fewer words.

\begin{lemma}[Junta size versus learning dimension]\label{lem:dimension}
For all sufficiently large $n$ and every integer $N$ with $2^{2\ell}\le N\le2^n$, set
\begin{equation*}
 t=\left\lfloor\frac{\log N}{2\ell}\right\rfloor,
 \qquad J=\lceil n/t\rceil~,
\end{equation*}
and define $r,B,N_0,K$ as in~Eqs.~(\ref{eq:precision}), (\ref{eq:N0}), and~(\ref{eq:junta-size}).
Then $1\le t\le\lfloor n/2\rfloor$, $r\le\ell$, $N_0\le N$, and
\begin{equation}\label{eq:overview-tradeoff}
 K=\Theta\!\left(\frac{n\log n}{\log N}
        \log\!\left(\frac{n\log n}{\log N}\right)\right)~,
\end{equation}
where $\Theta(\cdot)$ hides absolute numerical constants.
Thus, Lemma~\ref{lem:junta} applies in dimension $N$ with the above $K$.
The map $\Phi$ depends only on $n,N$.
\end{lemma}
\begin{proof}
Since $\log N\ge2\ell$, we have $t=\lfloor\log N/(2\ell)\rfloor\ge1$.
Also, $\log N\le n$ and $\ell\ge1$ give $t=\lfloor\log N/(2\ell)\rfloor\le\lfloor n/2\rfloor$.
Thus $2\le J=\lceil n/t\rceil\le n$.
Hence, $64J\le64n\le q=n^2$ for all sufficiently large $n$, so $r=\lceil\log(64J)\rceil\le\lceil\log q\rceil=\ell$.

The definitions of $t$ and $\ell$ give $q^t\le2^{\ell t}\le\sqrt N$.
Moreover, $N\ge2^{2\ell}$ gives $\sqrt N\ge2^\ell\ge n^2$.
Since $\ell=O(\log n)$, we have $\ell(n+1)\le n^2\le\sqrt N$ for all sufficiently large $n$. Using $r\le\ell$ and $J\le n$, we conclude
\[
 N_0=r(1+Jq^t)
 \le\ell(1+n\sqrt N)
 \le\ell(n+1)\sqrt N\le N~.
\]

Finally, since $\ell=\Theta(\log n)$, the definition of $t$ gives $t=\Theta(\log N/\log n)$ and $J=\Theta(n\log n/\log N)$.
By the definition of $r$, $K=(J+1)r=\Theta(J\log J)$.
Substituting the expression for $J$ gives~\eqref{eq:overview-tradeoff}.
\end{proof}

\begin{corollary}[Realizable examples]\label{cor:realizable}
Let $N$ satisfy the conditions of Lemma~\ref{lem:dimension}, and let $\Phi$ and $K$ be as in that lemma, with $K$ satisfying~\eqref{eq:overview-tradeoff}.
For every generating triple $(A,s,e)$ with $e\in\mathcal G_n$ and $b=As+e\pmod q$, the example oracle
\begin{equation}\label{eq:example}
 \rho\leftarrow\bits^M,\quad \mu\leftarrow\bits,\qquad
 (X,Y)=\bigl(\Phi(\Enc_{A,b}(\mu;\rho)),\mu\bigr)
\end{equation}
returns i.i.d.\ examples labeled by a fixed $K$-junta $f_s$ on $\bits^N$.
Given $n,N,A,b$, each call can be implemented in $N\poly(n)$ time, using $M+1$ independent random bits.
\end{corollary}
\begin{proof}
Lemma~\ref{lem:dimension} and~\eqref{eq:junta-dec} give $Y=f_s(X)$ for every choice of the coins. Encryption uses exact modular arithmetic, and Lemma~\ref{lem:junta} bounds the time to compute the feature map by $N\poly(n)$. These operations require neither $s$ nor $e$.
\end{proof}

\subsection{Quantitative security of Regev encryption}
\label{sec:encryption-security}\label{app:security}

We use the statistical hiding argument of \citet{Reg09} to convert a plaintext prediction advantage into an LWE distinguishing advantage.
A \emph{plaintext predictor} is a randomized algorithm $\mathcal P(A,b,c)$ that outputs one bit. Here $(A,b)$ is its public key and $c\in\Z_q^{n+1}$ is a challenge ciphertext.
Let $T(n)$ bound its running time on every well-formed public key and ciphertext and every choice of its internal coins. Thus, this bound also holds when the public key is uniform. The predictor may generate any number of additional encryptions using the public key within this time bound.

Define its success probability under LWE keys by
\begin{equation*}
 p_{\mathcal P}(n)=
 \Pr\!\left[
 \mathcal P(A,b,\Enc_{A,b}(\mu;\rho))=\mu
 \right]~,
\end{equation*}
where $(A,b)\sim\cL_n$, $\mu\leftarrow\bits$, and $\rho\leftarrow\bits^M$ are independent. The probability also includes the predictor's independent internal coins.
For an LWE distinguisher $\mathcal D$, write
\[
 \operatorname{Adv}(\mathcal D;n)=
 \left|\Pr_{\cL_n}[\mathcal D(A,b)=1]
       -\Pr_{\cU_n}[\mathcal D(A,b)=1]\right|~.
\]

\begin{lemma}[Quantitative encryption security]
\label{lem:encryption-security}
For every plaintext predictor $\mathcal P$ with time bound $T(n)$, there is an LWE distinguisher $\mathcal D_{\mathcal P}$ using only the supplied $(A,b)$ such that, for every $n$,
\begin{equation}\label{eq:security-interface}
 \operatorname{Adv}(\mathcal D_{\mathcal P};n)
 \ge p_{\mathcal P}(n)-\frac12-\xi_n,
 \qquad \xi_n=2^{-n-1}~.
\end{equation}
Its worst-case running time is at most $T(n)+\poly(n)$.
\end{lemma}

We first establish the statistical bound used in the proof.
Let $W=(A\mid b)$. Under a uniform public key, $W\leftarrow\Z_q^{M\times(n+1)}$ and an encryption of $0$ is $W^\top\rho$. We show that $(W,W^\top\rho)$ is close to $(W,V)$, where $V$ is uniform and independent of $W$.

For distributions $P,Q$ on a finite or countable set, define their total variation distance by
\[
 \TV(P,Q)=\frac12\sum_x|P(x)-Q(x)|~.
\]
For random variables $X,Y$, we also write $\TV(X,Y)$ for the total variation distance between their distributions.
We use two elementary properties: every event $E$ satisfies $|P(E)-Q(E)|\le\TV(P,Q)$, and applying the same randomized map to both distributions cannot increase total variation, regardless of the map's computation time.

\begin{lemma}[Leftover hash lemma, specialized]\label{lem:hash}
Let $M\ge1$ and $d\ge1$ be integers.
Let $W\leftarrow\Z_q^{M\times d}$, $R\leftarrow\bits^M$, and $V\leftarrow\Z_q^d$ be independent.
Then
\begin{equation}\label{eq:hash-general}
 \TV\bigl((W,W^\top R),(W,V)\bigr)
 \le\frac12\sqrt{\frac{q^d}{2^M}}~.
\end{equation}
In particular, for $d=n+1$ and the choices of $q$, $\ell$, and $M$ in \eqref{eq:parameters}, the right-hand side is at most
\[
 \xi_n=2^{-n-1}~.
\]
\end{lemma}
\begin{proof}
We use the following subset-sum form of the leftover hash lemma, as stated by \citet[Claim 5.3]{Reg09}.
Let $G$ be a finite abelian group and $M\ge1$ an integer.
Choose $g_1,\ldots,g_M$ independently and uniformly from $G$.
For each fixed $g=(g_1,\ldots,g_M)$, let $p_g$ be the distribution of $\sum_{i=1}^M R_i g_i$ for uniform $R\in\bits^M$. Then
\begin{equation}\label{eq:subset-sum-hash}
 \E_g\TV(p_g,\Unif(G))
 \le\frac12\sqrt{\frac{|G|}{2^M}}~.
\end{equation}
Note that \citet{Reg09} calls the full $L_1$ distance statistical distance, so the factor $1/2$ converts this bound to our total variation convention.

Set $G=\Z_q^d$. For each fixed matrix $H\in\Z_q^{M\times d}$, define
\[
 p_H(a)=\Pr_R[H^\top R=a],\qquad a\in\Z_q^d~.
\]
The rows of $W$ are independent uniform elements of $\Z_q^d$, so \eqref{eq:subset-sum-hash} applies to $p_W$. To compare the joint distributions of the matrix and vector, write $\pi(H)=\Pr[W=H]$. Both joint distributions have this same matrix marginal, and conditional on $W=H$, their second components have distributions $p_H$ and $\Unif(\Z_q^d)$, respectively. Therefore,
\begin{align*}
 &\TV\bigl((W,W^\top R),(W,V)\bigr)\\
 &\quad=\frac12\sum_{H\in\Z_q^{M\times d}}\sum_{a\in\Z_q^d}
       \pi(H)|p_H(a)-q^{-d}|\\
 &\quad=\E_W\TV(p_W,\Unif(\Z_q^d))\\
 &\quad\le\frac12\sqrt{\frac{q^d}{2^M}}~,
\end{align*}
which proves~\eqref{eq:hash-general}.

For the parameters used in the encryption scheme, $d=n+1$ and $M=(n+1)\ell+2n$, with $q\le2^\ell$. Therefore
\[
 \frac12\sqrt{\frac{q^{n+1}}{2^M}}
 \le\frac12\sqrt{\frac{2^{(n+1)\ell}}{2^{(n+1)\ell+2n}}}
 =2^{-n-1}~.
\]
\end{proof}

\begin{proof}[Proof of Lemma~\ref{lem:encryption-security}]
On input $(A,b)$, the distinguisher $\mathcal D_{\mathcal P}$ draws independent $\mu\leftarrow\bits$ and $\rho\leftarrow\bits^M$, forms $c=\Enc_{A,b}(\mu;\rho)$, and accepts if and only if
\[
 \mathcal P(A,b,c)=\mu~.
\]
Its running time is at most $T(n)+\poly(n)$. Under LWE keys,
\begin{equation}\label{eq:app-real}
 \Pr_{\cL_n}[\mathcal D_{\mathcal P}=1]=p_{\mathcal P}(n)~.
\end{equation}

Let $W=(A\mid b)$ be uniform on $\Z_q^{M\times(n+1)}$.
Lemma~\ref{lem:hash} bounds the total variation distance between $(W,W^\top\rho)$ and $(W,V)$ by $\xi_n$, where $V$ is independent of $W,\rho$ and uniform on $\Z_q^{n+1}$. The distinguisher samples an independent fair bit $\mu$ and shifts the last coordinate of $W^\top\rho$ by $\mu c_q$, producing an encryption of $\mu$.
For comparison, applying the same transformation to $V$ leaves it uniform and independent of $W,\mu,\rho$. Since this randomized transformation cannot increase total variation distance, we obtain
\begin{equation*}
 \TV\!\left(
 (W,\mu,\Enc_{A,b}(\mu;\rho)),
 (W,\mu,V)
 \right)\le\xi_n~.
\end{equation*}
In the right-hand experiment, $\mu$ is independent of the predictor's inputs and internal coins, so its prediction succeeds with probability $1/2$. By the total variation bound, the two success probabilities differ by at most $\xi_n$, so
\begin{equation}\label{eq:app-uniform}
 \Pr_{\cU_n}[\mathcal D_{\mathcal P}=1]\le\frac12+\xi_n~.
\end{equation}
Subtracting~\eqref{eq:app-uniform} from~\eqref{eq:app-real}, and bounding the signed difference by its absolute value, proves \eqref{eq:security-interface}.
\end{proof}

\subsection{From learning to LWE distinguishing}
\label{sec:reduction}\label{sec:learning-reduction}

In the following lemma, our learner uses the examples from Corollary~\ref{cor:realizable} to predict a fresh plaintext. Lemma~\ref{lem:encryption-security} then gives an LWE distinguisher.
We write $K(n,N)$ for $K$ from Lemma~\ref{lem:dimension}.

\begin{lemma}[Learning-to-LWE reduction]
\label{lem:reduction}\label{lem:runtime-reduction}
Suppose a learner learns $\kappa(N)$-juntas on $N$ variables in time $T(N)$ for all sufficiently large $N$.
Fix an algorithm that computes an integer $N(n)$ in time at most $C_N(n)$, with $2^{2\ell}\le N(n)\le2^n$ and $K(n,N(n))\le\kappa(N(n))$ for all sufficiently large $n$.
Then, there is an LWE distinguisher $\mathcal R_N(A,b)$ such that for all sufficiently large $n$
\begin{equation}\label{eq:advantage}
 \operatorname{Adv}(\mathcal R_N;n)
 \ge\frac{3}{10}~,
\end{equation}
and its worst-case running time satisfies
\begin{equation}\label{eq:arbitrary-runtime}
 T_{\mathcal R_N}(n)
 \le C_N(n)+N(n)T(N(n))\poly(n)~.
\end{equation}
\end{lemma}
\begin{proof}
Fix the learner $\mathcal A$, evaluator $\Eval$, and program for $N$.

\paragraph{Constructing the predictor.}
On input $(A,b,c^*)$, the predictor $\mathcal P_N$ returns $0$ if $n<n_0$, where $n_0$ is a fixed sufficiently large constant.
Otherwise, it computes $N=N(n)$ and runs $\mathcal A$ at dimension $N$ with $\varepsilon=\delta=1/10$, answering each example request using the example oracle from Corollary~\ref{cor:realizable} (\eqref{eq:example}).
The learner does not inspect $c^*$. Once it returns a hypothesis description $h$, the predictor outputs $\Eval(h,\Phi(c^*))$.

\paragraph{Prediction success and distinguishing advantage.}
Fix a sufficiently large $n$ for which $K(n,N(n))\le\kappa(N(n))$.
Conditional on any generating triple $(A,s,e)$ with $e\in\mathcal G_n$, Corollary~\ref{cor:realizable} gives independent examples from one fixed target in the learner's class. With probability at least $9/10$, the returned hypothesis has error at most $1/10$. The fresh encoded challenge is independent of the learner's execution and has the same example distribution, so prediction succeeds with probability at least $(9/10)^2=81/100$. By Lemma~\ref{lem:correctness}, the good event $e\in\mathcal G_n$ has probability at least $1-\gamma_n$.
The prediction success probability therefore satisfies
\begin{equation*}
 p_{\mathcal P_N}(n)\ge\frac{81}{100}(1-\gamma_n)~.
\end{equation*}
Applying Lemma~\ref{lem:encryption-security} to $\mathcal P_N$ gives a distinguisher $\mathcal R_N$ with
\[
 \operatorname{Adv}(\mathcal R_N;n)
 \ge p_{\mathcal P_N}(n)-\frac12-\xi_n
 \ge\frac{31}{100}-\frac{81}{100}\gamma_n-\xi_n~.
\]
Since $\gamma_n,\xi_n\to0$ as $n\to\infty$, the RHS is at least $3/10$ for all sufficiently large $n$, proving~\eqref{eq:advantage}.

\paragraph{Running time.}
Computing $N=N(n)$ takes at most $C_N(n)$ time.
The learner makes at most $T(N)$ example requests, each of which takes $N\poly(n)$ time by Corollary~\ref{cor:realizable}.
Computing the challenge feature vector $\Phi(c^*)$ takes $N\poly(n)$ time by Lemma~\ref{lem:junta}. Learning and evaluating the returned hypothesis take $O(T(N))$ time in total. Thus, the predictor runs in time $C_N(n)+N T(N)\poly(n)$.
Lemma~\ref{lem:encryption-security} adds only $\poly(n)$ time to obtain the distinguisher. This gives~\eqref{eq:arbitrary-runtime}.
\end{proof}

\paragraph{Choosing the learning dimension.}
We use two dimension choices. When $\kappa$ is computable, searching over integer dimensions gives both the polynomial-time result under Assumption~\ref{ass:no-subexp} and the quantitative results.
The fixed-$\beta$ polynomial-time result (under Assumption~\ref{ass:lwe}) uses an explicit choice, and does not require computability of $\kappa$.

\subsection{Polynomial-time hardness under Assumption~\ref{ass:lwe}}
\label{sec:proof-beta-poly}

\subsubsection{Choosing \texorpdfstring{$N(n)$}{N(n)}}

The following dimension choice ensures $K(n,N(n))\le\kappa(N(n))$ for all sufficiently large $n$ when $\kappa(N)=\Omega((\log N)^{1/\beta-1}\,(\log\log N)^2)$.

\begin{lemma}[An explicit feasible dimension]
\label{lem:fixed-beta-dimension}
Fix a rational $\beta\in(1/2,1)$ and $\eta>0$. There is a fixed integer $c\ge1$ such that $N(n)=2^{\lfloor cn^\beta\rfloor}$ lies in $[2^{2\ell},2^n]$ for all sufficiently large $n$ and satisfies
\[
 K(n,N(n))\le\eta(\log N(n))^{\frac{1}{\beta}-1}\,(\log\log N(n))^2~.
\]
\end{lemma}
\begin{proof}
Fix an integer $c\ge1$, to be chosen below, and set $N(n)=2^{\lfloor cn^\beta\rfloor}$. Since $\ell=\Theta(\log n)$ and $0<\beta<1$, for all sufficiently large $n$ we have $2^{2\ell}\le N(n)\le2^n$ and
\[
 \log N(n)\ge\frac c2 n^\beta,
 \qquad
 \log\log N(n)\ge\frac\beta2\log n~.
\]
Using~\eqref{eq:overview-tradeoff} and $\log(n\log n/\log N(n))\le2\log n$, we obtain
\[
 \frac{K(n,N(n))}
 { (\log N(n))^{1/\beta-1}(\log\log N(n))^2}
 \le\frac{A n(\log n)^2}
 { (\log N(n))^{1/\beta}(\log\log N(n))^2}
 \le C_\beta c^{-1/\beta}~,
\]
where $A,C_\beta>0$ are independent of $c$. Choosing a fixed integer $c$ large enough that $C_\beta c^{-1/\beta}\le\eta$ proves the claim.
\end{proof}

\subsubsection{Proof of Theorem~\ref{thm:beta-main}\textnormal{(i)}}
Suppose there is a learner with polynomial running time $T(N)$.
Choose $\eta>0$ so that
\[
 \kappa(N)\ge\eta(\log N)^{\frac{1}{\beta}-1}\,(\log\log N)^2
\]
eventually, and take $N(n)$ from Lemma~\ref{lem:fixed-beta-dimension}.
Then $K(n,N(n))\le\kappa(N(n))$ and $N(n) \in [2^{2\ell},2^n]$ eventually, and since $\beta$ is rational we have $C_N(n)=\poly(n)$. The program uses one fixed integer $c$ and never evaluates~$\kappa$.
Lemma~\ref{lem:reduction} gives a distinguisher with
\[
 T_{\mathcal R_N}(n)
 \le C_N(n)+N(n)T(N(n))\poly(n)
 \le\poly(n,N(n))=2^{O(n^\beta)}~.
\]
The distinguisher has advantage greater than $1/10$ at every sufficiently large $n$. This contradicts Assumption~\ref{ass:lwe}.
\qed

\subsection{Polynomial-time hardness under Assumption~\ref{ass:no-subexp}}
\label{sec:proof-main-poly}

\subsubsection{Searching for \texorpdfstring{$N(n)$}{N(n)}}

We search increasing integer dimensions until finding $N$ such that $K(n,N)\le\kappa(N)$.
The following lemma shows that the first such dimension $N(n)$ satisfies $N(n)=2^{o(n)}$ and can be found in $\poly(n,N(n))$ time.

\begin{lemma}[Searching for a feasible learning dimension]
\label{lem:dimension-search}
Let $\kappa$ be efficiently computable and satisfy
\[
 \kappa(N)=\omega\bigl(\log\log N\,\log\log\log N\bigr)~.
\]
Let $K(n,N)$ be from Lemma~\ref{lem:dimension}.
For all sufficiently large $n$, define $N(n)$ as the smallest integer $N\ge2^{2\ell}$ satisfying
\begin{equation}\label{eq:no-subexp-feasible}
 K(n,N)\le\kappa(N)~.
\end{equation}
Then, the search succeeds in time $C_N(n)\le\poly(n,N(n))$, and
\[
 2^{2\ell}\le N(n)\le2^n,\qquad \log N(n)=o(n)~.
\]
If, in addition, for a fixed rational $\beta\in(1/2,1)$,
\[
 \kappa(N)=\Omega\bigl((\log N)^{\frac{1}{\beta}-1}\,(\log\log N)^2\bigr)~,
\]
then $\log N(n)=O(n^\beta)$.
\end{lemma}
\begin{proof}
Test the integers $N\ge2^{2\ell}$ in increasing order and return the first one satisfying~\eqref{eq:no-subexp-feasible}.
Each test takes $\poly(n,N)$ time: we compute $\kappa(N)$ and $K$ from Lemma~\ref{lem:dimension}, and then check~\eqref{eq:no-subexp-feasible}.
A successful search examines at most $N(n)$ candidates, so its running time is at most $\poly(n,N(n))$.

To show that the search succeeds, fix $c\in(0,1)$ and set $N_c=2^{\lfloor cn\rfloor}$. For all sufficiently large $n$, this candidate lies in $[2^{2\ell},2^n]$. Since $\log\log N_c=\Theta(\log n)$ and $\log\log\log N_c=\Theta(\log\log n)$, \eqref{eq:overview-tradeoff} and the growth condition on $\kappa$ give
\[
 K(n,N_c)=O_c\bigl(\log n\,\log\log n\bigr),
 \qquad
 \kappa(N_c)=\omega\bigl(\log n\,\log\log n\bigr)~.
\]
Thus $N_c$ satisfies~\eqref{eq:no-subexp-feasible} for all sufficiently large $n$, so the search succeeds and $N(n)\le N_c$.
Since this holds for every fixed $c\in(0,1)$, we obtain $\log N(n)=o(n)$.

Under the additional growth condition, choose $\eta>0$ such that $\kappa(N)\ge\eta(\log N)^{1/\beta-1}(\log\log N)^2$ for all sufficiently large $N$.
Lemma~\ref{lem:fixed-beta-dimension} gives a fixed integer $c\ge1$ for which $N_{c,\beta}=2^{\lfloor cn^\beta\rfloor}$ satisfies~\eqref{eq:no-subexp-feasible} eventually. The search therefore returns $N(n)\le N_{c,\beta}$, proving $\log N(n)=O(n^\beta)$.
\end{proof}

\subsubsection{Proof of Theorem~\ref{thm:main}\textnormal{(i)}}
Suppose that $\kappa(N)$-juntas on $N$ variables are learnable in polynomial time $T(N)$. Apply Lemma~\ref{lem:dimension-search} with $\kappa(N)$ and set $N=N(n)$.
Thus, for every sufficiently large $n$, we have $2^{2\ell}\le N\le2^n$, $K(n,N)\le\kappa(N)$, and $\log N(n)=o(n)$.
The search finds $N(n)$ in time $C_N(n)\le\poly(n,N)$.
Lemma~\ref{lem:reduction} therefore gives a distinguisher whose worst-case running time satisfies
\[
 \begin{aligned}
 T_{\mathcal R_N}(n)
 &\le C_N(n)+N T(N)\poly(n)\\
 &\le\poly(n,N)
  =2^{o(n)}~.
 \end{aligned}
\]
The distinguisher's advantage is greater than $1/10$ at every sufficiently large $n$. This contradicts Assumption~\ref{ass:no-subexp}.
\qed

\subsection{Explicit time lower bounds}
\label{sec:proof-explicit-time}

For the quantitative bounds, we show that the first feasible dimension from Lemma~\ref{lem:dimension-search} satisfies $K=\Theta(\kappa(N))$.

\begin{lemma}[Junta size at the first feasible dimension]
\label{lem:quantitative-calibration}
Let $\kappa(N)$ be efficiently computable and satisfy
\[
 \kappa(N)=\omega\bigl(\log\log N\,\log\log\log N\bigr)~.
\]
Suppose also that $\kappa(N)=O(\log N)$ and there is a constant $C_\kappa\ge1$ such that, for all sufficiently large $N$,
\[
 \kappa(N+1)\le C_\kappa\kappa(N)~.
\]
Let $N=N(n)$ be the dimension from Lemma~\ref{lem:dimension-search}.
For all sufficiently large $n$,
\begin{equation}\label{eq:runtime-parameters}
 K(n,N)=\Theta(\kappa(N))~.
\end{equation}
Moreover,
\begin{equation}\label{eq:runtime-inversion}
 n=\Theta\!\left(
       \frac{\kappa(N)\log N}{\log\kappa(N)\,\log\log N}\right)~.
\end{equation}
\end{lemma}
\begin{proof}
Fix a sufficiently large $n$ and write $N=N(n)$.
At the initial candidate $2^{2\ell}$, \eqref{eq:overview-tradeoff} gives $K(n,2^{2\ell})=\Theta(n\log n)$, whereas $\kappa(2^{2\ell})=O(\ell)=O(\log n)$.
The initial candidate is therefore infeasible, and $N>2^{2\ell}$.
By minimality, $\kappa(N-1)<K(n,N-1)$.

Both $N-1$ and $N$ lie in $[2^{2\ell},2^n]$, and $\log(N-1)=\Theta(\log N)$. By~\eqref{eq:overview-tradeoff}, $K(n,N-1)=\Theta(K(n,N))$.
Using the assumption $\kappa(N)\le C_\kappa\kappa(N-1)$, we obtain
\[
 K(n,N)\le\kappa(N)\le C_\kappa\kappa(N-1)
        <C_\kappa K(n,N-1)=O(K(n,N))~.
\]
This proves~\eqref{eq:runtime-parameters}.

Let $x=n\log n/\log N$.
Eqs.~(\ref{eq:overview-tradeoff}) and~(\ref{eq:runtime-parameters}) give $\kappa(N)=\Theta(x\log x)$.
Taking logarithms yields $\log\kappa(N)=\Theta(\log x)$, so
\[
 x=\Theta\!\left(\frac{\kappa(N)}{\log\kappa(N)}\right)~.
\]
Since $\kappa(N)=O(\log N)$, we have $n\le n\log n=x\log N=O((\log N)^2)$.
Combining this with $\log N\le n$ and taking logarithms gives
\[
 \log\log N\le\log n\le2\log\log N+O(1)~.
\]
Thus $\log n=\Theta(\log\log N)$. Therefore,
\[
 n=\frac{x\log N}{\log n}
  =\Theta\!\left(
       \frac{\kappa(N)\log N}{\log\kappa(N)\,\log\log N}\right)~,
\]
which proves~\eqref{eq:runtime-inversion}.
\end{proof}

\subsubsection{Proof of Theorems~\ref{thm:main}\textnormal{(ii)} and~\ref{thm:beta-main}\textnormal{(ii)}}
Suppose that $\kappa(N)$ satisfies the hypotheses of either theorem and that $\kappa(N)$-juntas on $N$ variables are learnable in the corresponding time $T(N)$.
In both cases, Lemma~\ref{lem:dimension-search} provides a dimension $N=N(n)$ with $2^{2\ell}\le N\le2^n$ and $K(n,N)\le\kappa(N)$, computable in time $C_N(n)\le\poly(n,N)$.
Lemma~\ref{lem:quantitative-calibration} then gives~\eqref{eq:runtime-inversion}.

Lemma~\ref{lem:reduction} gives an LWE distinguisher with advantage at least $3/10$ for every sufficiently large $n$.
Using $C_N(n)\le\poly(n,N)$ and $n\le N$, its worst-case running time satisfies
\[
 \begin{aligned}
 T_{\mathcal R_N}(n)
 &\le C_N(n)+N T(N)\poly(n)\\
 &\le\poly(n,N)T(N)\le N^{O(1)}T(N)~.
 \end{aligned}
\]
Taking logarithms gives
\begin{equation}\label{eq:runtime-log-cost}
 \log T_{\mathcal R_N}(n)
 \le\log T(N)+O(\log N)~.
\end{equation}

For Theorem~\ref{thm:main}\textnormal{(ii)}, \eqref{eq:runtime-no-subexp} and~\eqref{eq:runtime-inversion} give $\log T(N)=o(n)$, while Lemma~\ref{lem:dimension-search} gives $\log N=o(n)$. Thus~\eqref{eq:runtime-log-cost} yields $T_{\mathcal R_N}(n)=2^{o(n)}$, contradicting Assumption~\ref{ass:no-subexp}.

For Theorem~\ref{thm:beta-main}\textnormal{(ii)}, \eqref{eq:runtime-beta} and~\eqref{eq:runtime-inversion} give $\log T(N)=O(n^\beta)$, while Lemma~\ref{lem:dimension-search} gives $\log N=O(n^\beta)$. Thus~\eqref{eq:runtime-log-cost} yields $T_{\mathcal R_N}(n)=2^{O(n^\beta)}$, contradicting Assumption~\ref{ass:lwe}.
\qed

\clearpage

\section{Additional corollaries and proofs}
\label{sec:scope}

\subsection{Decision trees}
\label{app:small-trees}

Below is the counterpart of Corollary~\ref{cor:tree-runtime} under Assumption~\ref{ass:lwe}.

\begin{corollary}[Quantitative decision-trees hardness under Assumption~\ref{ass:lwe}]
\label{cor:tree-runtime-beta}
Under Assumption~\ref{ass:lwe} for some fixed rational $\beta\in(1/2,1)$, if $\kappa(N)$ satisfies the conditions of Theorem~\ref{thm:beta-main}\textnormal{(ii)}, then decision trees with at most $2^{\kappa(N)}$ leaves are not learnable in time
\[
 T(N)=2^{O\!\left(\left[
 \frac{\kappa(N)\log N}{\log \kappa(N)\,\log\log N}
 \right]^\beta\right)}~.
\]

\end{corollary}

In particular, taking $\kappa(N)=\lfloor\log N\rfloor$ rules out learning decision trees with at most $N$ leaves in time
\[
 2^{O((\log N/\log\log N)^{2\beta})}~.
\]

\subsection{Bounded-width DNFs and sparse halfspace intersections}

\subsubsection{Bounded-width DNFs}
\label{app:dnf}

Below is the counterpart of Corollary~\ref{cor:dnf} under Assumption~\ref{ass:lwe}.

\phantomsection
\begin{corollary}[Hardness for bounded-width DNFs under Assumption~\ref{ass:lwe}]
\label{cor:dnf-beta}
Under Assumption~\ref{ass:lwe} for some fixed rational $\beta\in(1/2,1)$, if $\kappa(N)=\Omega((\log N)^{1/\beta-1}\,(\log\log N)^2)$, then $\kappa(N)$-DNFs on $\bits^N$ with at most $2^{\kappa(N)}$ terms are not polynomial-time improperly PAC learnable.
If, in addition, $\kappa(N)$ satisfies the conditions of Theorem~\ref{thm:beta-main}\textnormal{(ii)}, then they are not learnable in time
\[
 T(N)=2^{O\!\left(\left[
 \frac{\kappa(N)\log N}{\log\kappa(N)\,\log\log N}
 \right]^\beta\right)}~.
\]
\end{corollary}

\subsubsection{Intersections of sparse halfspaces}
\label{app:sparse-halfspaces}

We first state the counterpart of Corollary~\ref{cor:sparse-halfspaces} under Assumption~\ref{ass:lwe}, and then give the representation and parameter choice that prove both corollaries.

\phantomsection
\begin{corollary}[{\textls[-20]{Hardness for intersections of sparse halfspaces under Assumption~\ref{ass:lwe}}}]
\label{cor:sparse-halfspaces-beta}
Under Assumption~\ref{ass:lwe} for a fixed rational $\beta\in(1/2,1)$, if $\kappa(N)=\Omega((\log N)^{1/\beta-1}\allowbreak\,(\log\log N)^2)$, then intersections of at most $2^{\kappa(N)}$ halfspaces on $\bits^N$, each of which is $\kappa(N)$-sparse, are not polynomial-time improperly PAC learnable.
If, in addition, $\kappa(N)$ satisfies the conditions of Theorem~\ref{thm:beta-main}\textnormal{(ii)}, then they are not learnable in time
\[
 T(N)=2^{O\!\left(\left[
 \frac{\kappa(N)\log N}{\log\kappa(N)\,\log\log N}
 \right]^\beta\right)}~.
\]
Both bounds hold even when every weight lies in $\{0,1\}$, every threshold equals $1$, and the union of the halfspaces' supports has size at most $\kappa(N)$.
\end{corollary}

\begin{lemma}[Literal encoding for halfspaces]
\label{lem:literal-halfspace-encoding}
Let $d\ge1$ and $\kappa\ge0$ be integers, and define the map
\[
 \Psi_d:\bits^d\longrightarrow\bits^{2d},
 \qquad \Psi_d(x)=(x,\mathbf1_d-x)~,
\]
where $\mathbf1_d$ is the all-ones vector. For every $\kappa$-junta $f$ on $\bits^d$, there is an intersection $H_f$ of at most $2^\kappa$ halfspaces on $\bits^{2d}$ such that
\[
 H_f(\Psi_d(x))=f(x)\qquad\text{for every }x\in\bits^d~.
\]
Every weight lies in $\{0,1\}$ and every threshold equals one.
Each halfspace has at most $\kappa$ nonzero weights, and the union of their supports has size at most $2\kappa$.
\end{lemma}
\begin{proof}
Write $f(x)=g(x_I)$ for some $I\subseteq[d]$ with $|I|\le\kappa$.
For each assignment $a\in\bits^I$ with $g(a)=0$, we construct a halfspace that accepts $\Psi_d(x)$ exactly when $x_I\ne a$.

All halfspaces have their supports contained in
\[
 S=I\cup\{d+i:i\in I\}~.
\]
For each falsifying assignment $a$, define $v_a\in\{0,1\}^{2d}$ by
\[
 (v_a)_i=1-a_i,\qquad (v_a)_{d+i}=a_i
 \qquad(i\in I)~,
\]
and set all remaining weights to zero. This selects one literal coordinate for each $i\in I$, so $v_a$ has $|I|\le\kappa$ nonzero weights. On an encoded input,
\[
 \langle v_a,\Psi_d(x)\rangle
 =\sum_{i\in I}\bigl((1-a_i)x_i+a_i(1-x_i)\bigr)
 =\sum_{i\in I}\onefunc[x_i\ne a_i]~.
\]
Consequently, $\langle v_a,\Psi_d(x)\rangle\ge1$ if and only if $x_I\ne a$. Intersect these halfspaces over the falsifying assignments of $g$. There are at most $2^{|I|}\le2^\kappa$ of them, and the result agrees with $f$ on every encoded input.
\end{proof}

\begin{proof}[Proof of Corollaries~\ref{cor:sparse-halfspaces} and~\ref{cor:sparse-halfspaces-beta}]
Assume that we have a learner for the halfspace intersection class with sparsity bound $\kappa(N)$. We will construct a learner for $\kappa_0(d)$-juntas in dimension $d$, where
\[
 \kappa_0(d)=\left\lfloor\frac{\kappa(2d)}2\right\rfloor~.
\]
Lemma~\ref{lem:literal-halfspace-encoding} represents every $\kappa_0(d)$-junta on $\bits^d$ by an intersection of halfspaces on $\bits^{2d}$ with at most $2^{\kappa_0(d)}\le2^{\kappa(2d)}$ halfspaces and a union of supports of size at most $2\kappa_0(d)\le\kappa(2d)$.
All weights lie in $\{0,1\}$ and all thresholds equal one, so the intersection $H_f$ satisfies every restriction in the corollaries.

\paragraph{The learning reduction.}
Given labeled examples $(x,f(x))$ from any distribution $\mathcal D$ on $\bits^d$, feed $(\Psi_d(x),f(x))$ to the learner for intersections of halfspaces.
These examples are realizable by $H_f$. If the learner returns $\widehat H$, return $x\mapsto\widehat H(\Psi_d(x))$.
Since $H_f(\Psi_d(x))=f(x)$, this preserves prediction error and confidence. Computing $\Psi_d$ takes $O(d)$ time, so a halfspace learner with runtime $T(2d)$ gives a junta learner with runtime
\[
 T_{\mathrm{junta}}(d)\le\poly(d)T(2d)~,
\]
including hypothesis evaluation.

\paragraph{Polynomial-time hardness.}
In both corollaries, the bound $\kappa_0$ satisfies the growth condition of the corresponding junta theorem.
Moreover, $\kappa_0$ is efficiently computable whenever $\kappa$ is.
A polynomial-time halfspace learner would thus contradict Theorem~\ref{thm:main}\textnormal{(i)} or Theorem~\ref{thm:beta-main}\textnormal{(i)}, respectively.

\paragraph{Quantitative hardness.}
We first verify the remaining conditions on $\kappa_0$. Since $\kappa(2d)=O(\log(2d))$, we have $\kappa_0(d)=O(\log d)$.
If $\kappa(N+1)\le C_\kappa\kappa(N)$ eventually, applying this inequality twice yields
\[
 \kappa_0(d+1)
 \le \frac{C_\kappa^2}{2}\kappa(2d)
 \le C_\kappa^2\bigl(\kappa_0(d)+1\bigr)
 \le2C_\kappa^2\kappa_0(d)
\]
for all sufficiently large $d$, since $\kappa_0(d)\to\infty$.
Together with the growth and computability properties above, this verifies all conditions of the quantitative junta theorems.

Finally, we verify that the reduction preserves both quantitative bounds. Set
\[
 L_d=\frac{\kappa_0(d)\log d}
          {\log\kappa_0(d)\,\log\log d}
 =\Theta\!\left(
 \frac{\kappa(2d)\log(2d)}
      {\log\kappa(2d)\,\log\log(2d)}\right)~,
\]
where the last equality follows from $\kappa_0(d)=\Theta(\kappa(2d))$. The simulation bound gives
\[
 \log T_{\mathrm{junta}}(d)\le\log T(2d)+O(\log d)~.
\]
Under the assumptions for the quantitative bound in Corollary~\ref{cor:sparse-halfspaces}, write $h=\kappa_0(d)/\log\log d$. The growth condition gives $h=\omega(\log\log\log d)$, and hence
\[
 \frac{\log d}{L_d}
 =\frac{\log\kappa_0(d)}{h}
 =\frac{\log h}{h}+\frac{\log\log\log d}{h}
 \xrightarrow{d\to\infty}0~.
\]
This gives $\log d=o(L_d)$.
Thus, a learner for intersections of halfspaces with $\log T(2d)=o(L_d)$ would yield a junta learner with $\log T_{\mathrm{junta}}(d)=o(L_d)$, contradicting Theorem~\ref{thm:main}\textnormal{(ii)}.

Under the assumptions for the quantitative bound in Corollary~\ref{cor:sparse-halfspaces-beta}, the growth condition gives $\kappa_0(d)\ge c(\log d)^{1/\beta-1}(\log\log d)^2$ for some $c>0$ and all sufficiently large $d$.
Using $\log\kappa_0(d)\le2\log\log d$ for all sufficiently large $d$, we obtain
\[
 L_d\ge\frac{\kappa_0(d)\log d}{2(\log\log d)^2}
 \ge \frac c2(\log d)^{1/\beta}~.
\]
Thus $\log d=O(L_d^\beta)$.
A learner for intersections of halfspaces with $\log T(2d)=O(L_d^\beta)$ would therefore yield a junta learner with $\log T_{\mathrm{junta}}(d)=O(L_d^\beta)$, contradicting Theorem~\ref{thm:beta-main}\textnormal{(ii)}.
\end{proof}

\subsection{Bounded-degree polynomial thresholds and \texorpdfstring{$\mathbb F_2$}{F2} polynomials}
\label{app:sparse-polynomials}

Every $\kappa$-junta has a polynomial representation of degree at most $\kappa$ with at most $2^\kappa$ monomials over either $\mathbb F_2$ or $\mathbb R$. This gives lower bounds for learning $\mathbb F_2$ polynomials and real polynomial threshold functions $x\mapsto\onefunc[p(x)\ge 1/2]$ over $\bits^N$.

\phantomsection
\begin{corollary}
\label{cor:sparse-polynomials}
{\normalfont\bfseries\raggedright (Hardness for bounded-degree polynomial thresholds and \mbox{$\mathbb F_2$ polynomials})\par}
\noindent
Consider either polynomials over $\mathbb F_2$ or real polynomial threshold functions on $\bits^N$. In each case, the polynomial has degree at most $\kappa(N)$ and at most $2^{\kappa(N)}$ monomials.
\begin{enumerate}[label=\textnormal{(\roman*)}]
\item Under Assumption~\ref{ass:no-subexp}, no polynomial-time learner exists if $\kappa(N)=\omega(\log\log N\allowbreak\,\log\log\log N)$ and is efficiently computable.
      If $\kappa(N)$ satisfies the conditions of Theorem~\ref{thm:main}\textnormal{(ii)}, then no learner runs in time $N^{o(\kappa(N)/(\log\kappa(N)\,\log\log N))}$.
\item Under Assumption~\ref{ass:lwe} for a fixed rational $\beta\in(1/2,1)$, no polynomial-time learner exists if $\kappa(N)=\Omega((\log N)^{1/\beta-1}\,(\log\log N)^2)$.
      If $\kappa(N)$ satisfies the conditions of Theorem~\ref{thm:beta-main}\textnormal{(ii)}, then no learner runs in time
      \[
        2^{O\!\left(\left[
        \frac{\kappa(N)\log N}{\log\kappa(N)\,\log\log N}
        \right]^\beta\right)}~.
      \]
\end{enumerate}
\end{corollary}

\begin{proof}
Let $f(x)=g(x_I)$ be a $\kappa$-junta with $I\subseteq[N]$ and $|I|=d\le\kappa$.
Consider the polynomial
\[
 p(x)=\sum_{\substack{a\in\bits^I\\g(a)=1}}
       \prod_{\substack{i\in I\\a_i=1}}x_i
       \prod_{\substack{i\in I\\a_i=0}}(1-x_i)~.
\]
Each summand is the indicator of $x_I=a$, so $p(x)=f(x)$ on the Boolean cube. After expansion, every monomial is indexed by a subset of $I$.
Hence $p$ has degree at most $d$ and at most $2^d$ monomials. Reducing its coefficients modulo two gives a polynomial over $\mathbb F_2$ computing $f$.
Over $\mathbb R$, thresholding $p$ at $1/2$ also gives $f$. Thus, both classes contain every $\kappa$-junta. The claim now follows from Theorems~\ref{thm:no-subexp} and~\ref{thm:beta-main}.
\end{proof}

Note that when $\kappa=O(\log N)$, the number of monomials is polynomially bounded in $N$. While junta hardness implies hardness for polynomials of bounded degree, as shown above, we note that for \emph{unbounded} degree and polynomially many monomials, previous hardness results for DNFs already imply hardness of learning real polynomial threshold functions~\citep{DSS16,DV21,ABCM25} and $\mathbb F_2$ polynomials~\citep{DV21}.

\paragraph{Comparison with the best known upper bound.}
Both classes can be learned in $N^{O(\kappa)}$ time by mapping each input to the vector of its monomials of degree at most $\kappa$. In these coordinates, Gaussian elimination over $\mathbb F_2$ produces a polynomial consistent with the sample. For a real polynomial threshold target, linear programming produces a consistent hypothesis. The sample size and running time are polynomial in the number of degree-at-most-$\kappa$ monomials.
Under the conditions of Corollary~\ref{cor:sparse-polynomials}\textnormal{(i)}, this $N^{O(\kappa)}$ upper bound and the quantitative lower bound differ by a factor of $\log\kappa\,\log\log N$ in the exponent of $N$.
For additional discussion on upper bounds for learning polynomial thresholds and $\mathbb F_2$ polynomials, see \citet{HS07}.

\end{document}